\documentclass[12pt,a4paper]{article}
\usepackage{bm}

\usepackage[scale=0.8, top=2cm, bottom=3.5 cm]{geometry}

\usepackage{mathtools}
\usepackage{mathrsfs}

\usepackage{booktabs}
\usepackage{caption}
\usepackage{subcaption}
\usepackage{overpic} % for placing text over figures

\usepackage{graphicx} % Required for inserting images
\usepackage{amsmath}
\usepackage{amssymb}
\usepackage{xcolor}
\usepackage{graphicx}
\usepackage{float} 
\usepackage{comment}
\usepackage{url}
\usepackage{breqn} % For line breaks in math mode

\usepackage{siunitx} % Optional: For \num{} formatting
\usepackage{etoolbox} % For robust command definitions

\usepackage{longtable}
\usepackage{pdflscape} % Add this in your preamble
\usepackage{ragged2e}   % For \RaggedRight in table cells
\usepackage{array}
\usepackage{tabularx}
\usepackage{multirow}
\usepackage{enumitem}
\usepackage{afterpage}
\usepackage[nolist]{acronym}
\robustify\bfseries % Make \bfseries robust for use in table headers

\usepackage[colorlinks=true,linkcolor=blue,citecolor=blue,urlcolor=blue]{hyperref}

\begin{acronym}
\acro{EUCLID}[EUCLID]{efficient unsupervised constitutive law identification and discovery}
\acro{CANN}[CANN]{constitutive artificial neural network}
\acro{ICNN}[ICNN]{input convex neural network}
\acro{PANN}[PANN]{physics-augmented neural network}
\acro{CKAN}[CKAN]{constitutive Kolmogorov–Arnold network}
\acro{LASSO}[LASSO]{least absolute shrinkage and selection operator}
\acro{LARS}[LARS]{least angle regression}
\acro{OMP}[OMP]{orthogonal matching pursuit}
\acro{CV}[CV]{cross validation}
\acro{AIC}[AIC]{Akaike information criterion}
\acro{BIC}[BIC]{Bayesian information criterion}
\acro{NN}[NN]{neural network}
\acro{ANN}[ANN]{artificial neural network}
\acro{SEF}[SEF]{strain energy density function}
\acro{UT}[UT]{uniaxial tension}
\acro{EBT}[EBT]{equibiaxial tension}
\acro{BT}[BT]{biaxial tension}
\acro{SS}[SS]{simple shear}
\acro{PS}[PS]{pure shear}
\acro{RMSE}[RMSE]{root mean squared error}
\acro{NMSE}[NMSE]{normalized mean squared error}
\acro{RMS}[RMS]{root-mean-square}
\acro{RSS}[RSS]{residual sum of squares}
\acro{KKT}[KKT]{Karush-Kuhn-Tucker}
\acro{OLS}[OLS]{ordinary least squares}
\end{acronym}

\newcommand{\transpose}{^{\top}}
\newcommand{\minusTranspose}{^{-\top}}

\long\def\symbolfootnote[#1]#2{\begingroup \def\thefootnote{\fnsymbol{footnote}}\footnote[#1]{#2} \endgroup} 

\renewcommand{\vec}[1]{ \ensuremath{ \mathbf{ #1 } } }
\newcommand{\ten}[1]{ \ensuremath{ \bm{ \mathsf{ #1 } } } }

\newcommand{\gten}[1]{ \ensuremath {\boldsymbol{#1} } }

\newcommand{\tenC}{\ensuremath{ \ten{C} }}

\newcommand{\tenF}{\ensuremath{ \ten{F} }}

\newcommand{\tenM}{\ensuremath{ \ten{M} }}

\newcommand{\tenP}{\ensuremath{ \ten{P} }}
\newcommand{\tenQ}{\ensuremath{ \ten{Q} }}

\newcommand{\veca}{\ensuremath{ \vec{a} }}

\newcommand{\vecc}{\ensuremath{ \vec{c} }}

\newcommand{\vecf}{\ensuremath{ \vec{f} }}

\newcommand{\vecn}{\ensuremath{ \vec{n} }}

\newcommand{\vecr}{\ensuremath{ \vec{r} }}
\newcommand{\vecs}{\ensuremath{ \vec{s} }}

\newcommand{\vecw}{\ensuremath{ \vec{w} }}

\newcommand{\vecP}{\ensuremath{ \vec{P} }}

\newcommand{\tenPsi       }{\ensuremath{ \gten{\Psi} }}

\renewcommand{\ln}[1]{\text{$\hspace{0.1cm}$ln$\left(#1\right)$}}
\renewcommand{\exp}[1]{\ensuremath{ \,\text{exp}{\left( #1 \right)} }}

\title{Automated Constitutive Model Discovery \\ by Pairing Sparse Regression Algorithms with Model Selection Criteria}

\author{Jorge-Humberto Urrea-Quintero\textsuperscript{1}\thanks{Corresponding author: \href{mailto:jorge.urrea-quintero@tu-braunschweig.de}{jorge.urrea-quintero@tu-braunschweig.de}}, David Anton\textsuperscript{1}, \\ Laura De Lorenzis\textsuperscript{2}, Henning Wessels\textsuperscript{1}}

\date{%
  \textsuperscript{1}Institute of Applied Mechanics, Division Data-Driven Modeling of Mechanical Systems, Technische Universität Braunschweig, Pockelsstr. 3, 38106 Braunschweig, Germany \\
  \textsuperscript{2}Computational Mechanics Group, Eidgenössische Technische Hochschule Zürich, Tannenstrasse 3, 8092 Zürich, Switzerland 
}

\begin{document}

\maketitle

\paragraph{Abstract}
The automated discovery of constitutive models from data has recently emerged as a promising alternative to the traditional model calibration paradigm. 
In this work, we present a fully automated framework for constitutive model discovery that systematically pairs three sparse regression algorithms (least absolute shrinkage and selection operator (LASSO), least angle regression (LARS), and orthogonal matching pursuit (OMP)) with three model selection criteria: $K$-fold cross-validation (CV), Akaike information criterion (AIC), and Bayesian information criterion (BIC). 
This pairing yields nine distinct algorithms for model discovery and enables a systematic exploration of the trade-off between sparsity, predictive performance, and computational cost. 
While LARS serves as an efficient path-based solver for the $\ell_1$-constrained problem, OMP is introduced as a tractable heuristic for $\ell_0$-regularized selection. 
The framework is applied to both isotropic and anisotropic hyperelasticity, utilizing both synthetic and experimental datasets. 
Results reveal that all nine algorithm-criterion combinations perform consistently well in discovering isotropic and anisotropic materials, yielding highly accurate constitutive models.
These findings broaden the range of viable discovery algorithms beyond $\ell_1$-based approaches such as LASSO.


\newpage

% \tableofcontents

% \newpage

\section{Introduction}

The formulation of constitutive laws is central to continuum mechanics and enables the predictive capability of numerical simulations in solid mechanics. In hyperelasticity, classical approaches rely on postulating a functional form for the \ac{SEF}, grounded in invariance principles, material symmetries, and thermodynamic arguments \cite{ogden_nonLinearElasticDeformations_1984,holzapfel_nonlinearSolidMechanics_2000}. The model parameters are then identified by fitting the model to experimental data, with the model structure remaining fixed throughout the calibration process \cite{steinmann_hyperelasticModelsTreloar_2012,ricker_systematicFittingHyperelasticity_2023}. While this paradigm has produced many successful models, its effectiveness depends critically on the adequacy of the initial functional assumption \cite{kirchdoerfer_dataDrivenComputationalMechanics_2016,gurtin_mechanicsThermodynamicsOfContinua_2010,holzapfel_nonlinearSolidMechanics_2000,ricker_systematicFittingHyperelasticity_2023}.

The increasing availability of high-resolution experimental measurements and the rise of computational resources have enabled a complementary paradigm: {data-driven model discovery}. In this new paradigm, rather than fixing the model structure {a priori}, one specifies a library of candidate terms derived from the kinematic invariants and admissible functional forms, and seeks a reduced subset that best describes the data while satisfying physical constraints.
The \ac{EUCLID} framework \cite{flaschel_unsupervisedDiscoveryEUCLID_2021,flaschel_brainEUCLID_2023} pioneered this approach for hyperelasticity by constructing libraries from classical Ogden and generalized Mooney–Rivlin terms, later extended to include features from, e.g., the Arruda–Boyce, Haines–Wilson, and Gent–Thomas models \cite{joshi_BayesianEUCLID_2022}. Extensions to inelastic materials have also been proposed, in the most general case by utilizing libraries derived from the framework of generalized standard materials \cite{flaschel_discoveringPlasticityModels_2022, marino_linearViscoelasticityEUCLID_2023, flaschel_generalizedStandardMaterialModelsEUCLID_2023, flaschel_convexNN_2025}. While the original \ac{EUCLID} formulation focused on unsupervised discovery from displacement data, the same libraries have also been used in supervised learning settings with stress–strain data \cite{flaschel_brainEUCLID_2023}.

\Acp{ANN} have also been explored for data-driven constitutive modeling of solids \cite{linka_CANNs_2021,linka_newCANNs_2023,fuhg_reviewDataDrivenConstitutiveLaws_2025,dornheim_reviewNNConstitutiveModeling_2024}. Early work focused on purely data-driven architectures for hyperelasticity \cite{shen_NNBasedConstitutiveModel_2004,liang_NNBasedConstitutiveModelElastomers_2008}, viscoelasticity \cite{alhaik_predictionNonlinearViscoelasticityANN_2006,jung_NNRateDependentMaterials_2006}, plasticity \cite{huang_MLBasedPlasticityModel_2020}, and viscoplasticity \cite{furukawa_implicitConstitutiveModeling_1998}. While such models offer great flexibility, the absence of embedded physical constraints can lead to nonphysical or unstable responses in finite element simulations \cite{fuhg_reviewDataDrivenConstitutiveLaws_2025}. Recent developments address this by incorporating thermodynamic and structural constraints directly into the network architecture \cite{klein_polyconvexAnisotropicHyperelasticityWithNN_2022,linden_NNHyperelasticityEnforcingPhysics_2023,tepole_polyconvexPANNs_2025,fuhg_hyperelasticAnisotropyTensorBasedNN_2022,kalina_NNMeetsAnisotropicHyperelasticity_2025,dammaß_whenInvariantsMatter_2025}, and by hybrid approaches such as \acp{CKAN} \cite{thakolkaran_inputConvexKANs_2025,abdolazizi_CKANs_2025}.
Extreme sparsification has also been proposed to shrink the set of active parameters in an \ac{ANN}-based architecture to the minimum possible, thereby gaining some interpretability of the constitutive model \cite{fuhg_extremeSparsification_2024}.

Despite their versatility, \ac{ANN}-based approaches often lack the interpretability and generalization properties afforded by closed-form constitutive models. In contrast, constructing the library from combinations of well-established generalized models, such as Mooney–Rivlin or Ogden forms, anchors the discovery process in physical principles while allowing sufficient flexibility to capture a broad range of non-linear responses \cite{flaschel_brainEUCLID_2023}.
This approach facilitates comparison with decades of experimental and modeling literature and allows for interpretability of the obtained constitutive model.

\Acp{CANN}, as introduced in, e.g., \cite{linka_newCANNs_2023}, are often presented as \ac{ANN}-based discovery tools. However, these \acp{CANN} are not arbitrary black-box models but are built from modular constitutive library terms embedded within a network architecture \cite{linka_newCANNs_2023, stpierre_principalStretchBasedCNN_2023,linka_brainCANNs_2023,peirlinck_universalMaterialSubroutineHyperelasticity_2024,tac_benchmarkingDataDrivenHyperelasticity_2024,martonová_modelDiscoveryCardiacTissue_2025,martonova_discoveringDispersion_2025}. In other words, \acp{CANN} structurally function as model libraries. Therefore, in the present work, we treat \acp{CANN} as such, compatible with model libraries based on Mooney–Rivlin and Ogden model terms.

Across the different model discovery frameworks, sparsity-promoting regression techniques have been the algorithmic backbone that enables systematic identification of compact models from large model libraries. This has been true in the context of equation discovery for dynamical systems \cite{brunton_discoveringGoverningEquations_2016,champion_sparseOptimizationPIModels_2020} and automated constitutive modeling \cite{flaschel_unsupervisedDiscoveryEUCLID_2021,flaschel_brainEUCLID_2023,flaschel_discoveringPlasticityModels_2022,flaschel_generalizedStandardMaterialModelsEUCLID_2023,mcculloch_LPRegularizationModelDiscovery_2024}.
In settings where the model library is formulated as a linear combination of many model terms, the \ac{LASSO} \cite{tibshirani_regressionShrinkageSelectionLasso_1996} is the most commonly adopted technique due to its convex formulation \cite{flaschel_unsupervisedDiscoveryEUCLID_2021,flaschel_brainEUCLID_2023,flaschel_discoveringPlasticityModels_2022,marino_linearViscoelasticityEUCLID_2023,flaschel_generalizedStandardMaterialModelsEUCLID_2023}. When the model library is non-linear with respect to its parameters, more general \(\ell_p\)-regularized formulations have been investigated \cite{flaschel_discoveringPlasticityModels_2022,mcculloch_LPRegularizationModelDiscovery_2024,martonová_modelDiscoveryCardiacTissue_2025}.

While sparse regression provides a systematic way to generate a set of candidate models with varying model sizes, it does not in itself prescribe which model from this set is the most appropriate one for the specific problem at hand. For instance, the determination of the final model has often been performed by the user, e.g., by visual inspection of the error–complexity Pareto front \cite{flaschel_unsupervisedDiscoveryEUCLID_2021,flaschel_brainEUCLID_2023,martonová_modelDiscoveryCardiacTissue_2025}, {although automated strategies for selecting the regularization parameter have also been proposed, such as selecting the sparsest model below a predefined cost threshold \cite{flaschel_generalizedStandardMaterialModelsEUCLID_2023}.}
However, in the context of statistical learning, this final choice is commonly governed by model selection criteria \cite{hastie_statisticalLearning_2009}, which evaluate the trade-off between predictive accuracy and model complexity. Commonly used criteria include information-theoretic measures such as the \ac{AIC} \cite{akaike_AkaikeInformationCriterion_2011} and the \ac{BIC} \cite{schwarz_estimatingDimensionOfModel_1978,neath_BayesianInformationCriterion_2012}, as well as data-driven approaches like $K$-fold \ac{CV} \cite{bengio_varianceKFoldCV_2003,arlot_surveyCVModelSelection_2010}. Comparative studies \cite{burnham_modelSelection_2002,fan_overviewVariableSelection_2010,vrieze_differenceAICAndBIC_2012} have shown that no single criterion is universally optimal; rather, their relative performance depends on factors such as noise level, sample size, and the correlation structure within the library. For this reason, pairing sparse regression algorithms with different selection criteria can increase robustness to these factors, and it enables a more systematic exploration of the trade-offs inherent in automated model discovery.

From the literature on constitutive model discovery, two key gaps can be identified: \textbf{i.} The model discovery problem can be solved using different, more efficient, sparse regression algorithms beyond LASSO, and \textbf{ii.} model selection criteria have not been paired with sparse regression algorithms in this context, despite their widespread adoption in statistical learning for balancing model complexity and predictive performance \cite{hastie_statisticalLearning_2009,burnham_modelSelection_2002,arlot_surveyCVModelSelection_2010}.

The present work addresses these gaps by introducing a fully automated framework for constitutive model discovery that systematically combines three sparse regression algorithms, namely, LASSO, \ac{LARS} \cite{efron_leastAngleRegression_2004}, and \ac{OMP} \cite{pati_orthogonalMatchingPursuit_1993,mallat_matchingPursuits_1993}, with three model selection criteria: \ac{AIC} \cite{akaike_AkaikeInformationCriterion_2011}, \ac{BIC} \cite{neath_BayesianInformationCriterion_2012}, and $K$-fold \ac{CV} \cite{bengio_varianceKFoldCV_2003}. 

Path-based and greedy algorithms offer attractive alternatives for constructing sparse models from large candidate libraries. 
\ac{LARS} \cite{efron_leastAngleRegression_2004,khan_robustLinearModelSelectionLARS_2007,blatman_adaptiveSparsePCEBasedOnLARS_2011,zhang_LARSForModelSelection_2014} is a path-based algorithm that efficiently traces the entire solution path of the $\ell_1$-constrained optimization problem.
In contrast, \ac{OMP} \cite{pati_orthogonalMatchingPursuit_1993,tropp_greedAlgorithmsSparseApproximation_2004,blumensath_differenceOMPAndOrthogonalLeastSquares_2007} is a heuristic method that approximates the solution of the $\ell_0$-regularized problem. 
By adding at each iteration the term most strongly correlated with the current residual, \ac{OMP} circumvents the combinatorial complexity of exact $\ell_0$-minimization while still yielding compact models.
While not greedy in the strict sense, \ac{LARS} shares with \ac{OMP} the sequential, term-by-term inclusion strategy, initiating the model with a single active term and progressively expanding the active set. 
{In contrast, LASSO defines an $\ell_1$-penalized least-squares problem that promotes sparsity through coefficient shrinkage controlled by a regularization parameter. 
Following the interpretation in \cite{hastie_statisticalLearning_2009}, the LASSO solution path may be viewed either as a shrinkage from the fully populated parameter vector or, equivalently, as a forward inclusion process starting from the empty parameter vector.}
An advantage of \ac{LARS} and \ac{OMP} is that these algorithms expose the hierarchy of term relevance, which is particularly advantageous in model discovery frameworks.

{Recently, in \cite{linka_bestInClassModeling_2024}, a model discovery framework named {best--in--class} was introduced. This framework constructs constitutive models through a bottom-up densification strategy. Starting with the single term that provides the best initial fit, the {best--in--class} approach adds new terms by iteratively selecting the one whose inclusion results in the augmented model with the lowest fitting error. This process continues until a prescribed accuracy target is met, thereby preventing the combinatorial explosion that occurs with exhaustive subset selection. Conceptually, this strategy is closely related to greedy forward-selection algorithms such as \ac{OMP}, which are likewise employed in this work to discover sparse constitutive representations. Both frameworks share the underlying principle of incremental model expansion and sparsity control while differing in formulation: {best--in--class} evaluates the overall fitting performance of candidate models from a discrete library, whereas \ac{OMP} operates in a continuous regression setting, selecting terms based on correlation with the current residual followed by analytical orthogonalization.}

The key contributions of this work can be summarized as follows:
\begin{enumerate}
    \item {\textbf{Systematic integration and comparison of model selection criteria in constitutive model discovery.} To the best of our knowledge, this is the first study to systematically integrate and compare well-established model selection criteria (AIC, BIC, CV) with a suite of sparse regression algorithms (LASSO, LARS, OMP) in this context. This moves beyond user-based Pareto-front inspection and complements other previously proposed automated strategies, e.g., in \cite{flaschel_generalizedStandardMaterialModelsEUCLID_2023}.}
    
    \item \textbf{Extension beyond \ac{LASSO} to forward selection and greedy algorithms.} We acknowledge that, during the preparation of this manuscript, the preprint \cite{flaschel_LARSModelDiscovery_2025} was released, introducing LARS in the context of constitutive model discovery as a path-based algorithm for the $\ell_1$-constrained problem. In the present work, we also consider \ac{LARS}, but pair it with multiple model selection criteria, and extend the scope to \ac{OMP} as a tractable heuristic method for $\ell_0$-regularized selection. We believe that the dual inclusion of \ac{LARS} and \ac{OMP} broadens the algorithmic basis of constitutive model discovery and enables a more comprehensive assessment of the trade-offs between $\ell_1$ and $\ell_0$ formulations.

\end{enumerate}

The remainder of the paper is organized as follows. 
Section~\ref{sec:Methodology} introduces the proposed automated model discovery framework, detailing the three sparse regression algorithms (\ac{LASSO}, \ac{LARS}, and \ac{OMP}) and the model selection criteria (\ac{AIC}, \ac{BIC}, and \ac{CV}). 
Section~\ref{sec:model_libraries} describes the material model libraries, including the kinematic assumptions dictated by the deformation modes present in the datasets and the corresponding stress--strain relationships required for model evaluation. 
Section~\ref{sec:numerical_results} presents numerical experiments for isotropic and anisotropic hyperelasticity using both synthetic and experimental datasets.
Finally, Section~\ref{sec:conclusions} summarizes the main findings.

% \newpage

\section{Methodology}
\label{sec:Methodology}

In this section, we present the methodology for discovering material constitutive models. In Section~\ref{subsec:model_discovery_problem}, we formulate the problem. In Section~\ref{subsec:computational_approach}, we present a computational solution approach that pairs sparse regression algorithms with model selection criteria. 

\subsection{Problem Formulation}
\label{subsec:model_discovery_problem}

The central goal of this work is to discover a parsimonious and physically meaningful \ac{SEF}, $W$, that accurately describes the mechanical response of a hyperelastic material under different deformation modes, e.g., \ac{UT}, \ac{BT}, \ac{PS}, etc.. 
For an incompressible hyperelastic material, the \ac{SEF} takes the form:
\begin{equation}\label{eq:w_incompressible_split}
    W({\tenF}, p) = \overline{W}({\tenF}) - p(J-1), 
\end{equation}
where ${\tenF}$ is the deformation gradient, $\overline{W}$ is the isochoric \ac{SEF}, $p$ is the hydrostatic pressure, and $J = \operatorname{det} {\tenF} = 1$ is enforced. We postulate that the isochoric \ac{SEF} can be represented as a linear combination of $n_\phi$ basis functions, $\{\phi_{j}\}_{j=1}^{n_\phi}$:
\begin{equation}\label{eq:w_series_expansion}
    \overline{W}({\tenF}; \vecc, \vecw) = \sum_{j=1}^{n_\phi} c_{j} \phi_{j}({\tenF}; \vecw_{j}).
\end{equation}
Here, $\vecc \in \mathbb{R}^{n_\phi}$ is the vector of unknown linear coefficients, and $\vecw$ is a vector of non-linear parameters internal to the basis functions. 
The representation \eqref{eq:w_series_expansion} follows the standard library-based identification paradigm used in data-driven constitutive modeling and model discovery \cite{flaschel_unsupervisedDiscoveryEUCLID_2021,brunton_discoveringGoverningEquations_2016}.

\noindent Following the framework of continuum solid mechanics, the first Piola-Kirchhoff stress tensor $\ten{P}$ is derived from a scalar-valued \ac{SEF} $W$ as follows \cite{holzapfel_nonlinearSolidMechanics_2000,gurtin_mechanicsThermodynamicsOfContinua_2010}:
\begin{equation}\label{eq:pk_tensor_model}
    \tenP({\tenF}; \vecc, \vecw) = \sum_{j=1}^{n_{\phi}} c_{j}  \frac{\partial \phi_{j}({\tenF}; \vecw_{j})}{\partial \tenF} - p\, {\tenF}^{\minusTranspose}.
\end{equation}

\noindent \textbf{The constitutive model discovery problem} consists in finding the optimal set of sparse linear coefficients $\vecc$ and non-linear parameters $\vecw$ that best describe the material's response, given a model library as defined in Eq.~\eqref{eq:w_series_expansion} and the dataset $\mathcal{D} = \bigl\{\{\hat{\tenF}^{(k,d)}, \hat{\vecP}^{(k,d)}\}_{d=1}^{n_{d}^{(k)}}\bigr\}_{k=1}^{n_{k}}$. Here, $n_{k}$ denotes the total number of deformation modes considered in the mechanical test and $n_{d}^{(k)}$ is the number of stress-deformation pairs measured in the deformation mode $k$. The vector $\hat{\vecP}^{(k,d)}$ contains only the stress components of $\hat{\tenP}^{(k,d)}$ that are measured in the specific deformation mode $k$. All measured stresses for deformation mode $k$ are then stacked in one vector $\hat{\vecP}^{(k)} = \bigl[\hat{\vecP}^{(k,1)\transpose}, \cdots, \hat{\vecP}^{(k,n_{d}^{(k)})\transpose} \bigr]\transpose$. Note that the size of $\hat{\vecP}^{(k)}$ depends on both the number of measured stress-deformation pairs $n_{d}^{(k)}$ and the number of observed stress components for the deformation mode $k$. To simplify the notation in the following equations, we introduce the set $\hat{\mathcal{F}}^{(k)} = \bigl\{\hat{\tenF}^{(k,d)}\bigr\}_{d=1}^{n_{d}^{(k)}}$ which collects all deformation gradients for the deformation mode $k$.

Furthermore, $\tenP^{(k,d)}$ denotes the predicted stress tensor calculated for the deformation gradient $\hat{\tenF}^{(k,d)}$ from Eq.~\eqref{eq:pk_tensor_model}. Consequently, $\vecP^{(k,d)}$ is the vector that contains only the predicted stresses for the measured components and $\vecP^{(k)}$ comprises all $\vecP^{(k,d)}$ for the deformation mode $k$ in one vector. For each stress-deformation pair $(k,d)$, $p^{(k,d)}$ is the Lagrange multiplier which is determined by the boundary conditions of the specific deformation mode (e.g., a zero-stress condition on a free surface) \cite{holzapfel_nonlinearSolidMechanics_2000,bonet_nonlinearContinuumMechanicsFEM_1997}.

Ultimately, the goal is to solve the following optimization problem:

\noindent \textbf{Problem 1: non-smooth $\ell_0$-constrained constitutive model discovery}
\begin{equation} \label{eq:abstract_optimization_problem}
    \bigl\{\vecc^*_{n^{\mathcal{A}}_\phi}, \vecw^*_{n^{\mathcal{A}}_\phi}\bigr\} = 
    \arg\min_{\vecc, \vecw} \left\{ 
        \sum_{k=1}^{n_{k}} 
        \left\| 
            \hat{\mathbb{W}}^{(k)} \left( \hat{\vecP}^{(k)} 
            - \vecP^{(k)}(\hat{\mathcal{F}}^{(k)}; \vecc, \vecw) \right) 
        \right\|^2_2 
        \right\} 
    \quad \text{subject to} \quad \|\vecc\|_0 \le n^{\mathcal{A}}_\phi,
\end{equation}
where $\|\vecc\|_0$ is the $\ell_0$-pseudo-norm of $\vecc$ counting the non-zero coefficients, and $n^{\mathcal{A}}_\phi \ll n_\phi$ is the desired model complexity, i.e., the number of remaining active model terms.
Directly solving the $\ell_0$-constrained problem is NP-hard \cite{natarajan_sparseApproximateSolutions_1995}. That is, this problem is computationally intractable due to the combinatorial nature of the $\ell_0$-pseudo-norm and the non-linear dependence on $\vecw$. 

A common approach to formulate a tractable version of \textbf{Problem 1} that still promotes sparsity is to relax the non-convex $\ell_0$-pseudo-norm constraint by replacing it with a convex surrogate, the $\ell_1$-norm \cite{donoho_optimallySparseRepresentation_2003}. 
Considering the $\ell_1$-norm, \textbf{Problem 1} is therefore recast as: 

\noindent \textbf{Problem 2: relaxed $\ell_1$-constrained constitutive model discovery}
\begin{equation} \label{eq:l1_constrained_problem_abstract}
    \bigl\{\vecc^*_\tau, \vecw^*_\tau\bigr\} = 
    \arg\min_{\vecc, \vecw} \left\{ 
        \sum_{k=1}^{n_{k}} 
            \left\| 
                \hat{\mathbb{W}}^{(k)} \left( \hat{\vecP}^{(k)} 
                - \vecP^{(k)}(\hat{\mathcal{F}}^{(k)}; \vecc, \vecw) \right) 
            \right\|^2_2 
    \right\} 
    \quad \text{subject to} \quad \|\vecc\|_1 \le \tau,
\end{equation}
where $\|\vecc\|_1 = \sum_j |c_{j}|$ is the $\ell_1$-norm of $\vecc$ and $\tau \ge 0$ denotes the budget parameter controlling sparsity. A smaller value of $\tau$ enforces greater sparsity.

The objective function, as formulated in Eqs.~\eqref{eq:abstract_optimization_problem} and \eqref{eq:l1_constrained_problem_abstract}, corresponds to the sum of squared weighted residuals. $\hat{\mathbb{W}}^{(k)}$ is a matrix, typically diagonal.
In this work, we use a per-deformation mode scalar weight based on the characteristic \ac{RMS} stress:
\begin{equation}\label{eq:weight_matrix_definition}
\hat{\mathbb{W}}^{(k)} = \hat{\omega}^{(k)} \mathbb{I} = \left( \frac{\hat{P}_{\text{rms}}}{\hat{P}_{\text{rms}}^{(k)}} \right) \mathbb{I}.
\end{equation}
Here, $\mathbb{I}$ is the identity matrix, and $\hat{P}_{\text{rms}}^{(k)}$ is the \ac{RMS} of all stress observations $n_{\mathrm{obs}}^{(k)}$ within the vector ${\hat{\vecP}}^{(k)} \in \mathbb{R}^{n_{\mathrm{obs}}^{(k)}}$. The term $\hat{P}_{\text{rms}}$ is a global scaling factor, computed as the \ac{RMS} of all per-deformation mode $\hat{P}_{\text{rms}}^{(k)}$ values from $k=1, \dots, n_{k}$, which ensures that the weights are dimensionless and in a reasonable numerical range. That is:
\begin{equation}\label{eq:sigma_rms_k}
    \hat{P}_{\text{rms}}^{(k)} = \sqrt{\frac{1}{n_{\mathrm{obs}}^{(k)}} \sum_{i=1}^{n_{\mathrm{obs}}^{(k)}} \left( \hat{P}_{i}^{(k)} \right)^2},
    \qquad
    \hat{P}_{\text{rms}} = \sqrt{\frac{1}{n_{k}} \sum_{k=1}^{n_{k}} \left( \hat{P}_{\text{rms}}^{(k)} \right)^2}.
\end{equation}

This formulation normalizes the contribution of each deformation mode $k$ to the total loss. This is necessary because experimental data from different deformation modes may yield stress magnitudes, that differ in the order of magnitude. 
Alternative weighting strategies could also be employed, for instance, using the inverse of the measurement uncertainty for each data point, or more generally, using the inverse of the error covariance matrix to account for correlated errors \cite{seber_linearRegressionAnalysis_2003}.

While the $\ell_1$-norm makes the sparsity constraint convex, the objective remains non-linear and non-convex due to the presence of $\vecc$ and the non-linear parameters $\vecw$, leading to a bi-linear/non-linear inverse problem \cite{flaschel_unsupervisedDiscoveryEUCLID_2021}. 
Hence, a direct, simultaneous optimization with respect to both $\vecc$ and $\vecw$ can be computationally prohibitive and prone to poor local minima. 
We therefore require a numerical strategy to solve either the optimization problem \eqref{eq:abstract_optimization_problem} or \eqref{eq:l1_constrained_problem_abstract} efficiently.

\subsection{Computational Solution Approach}
\label{subsec:computational_approach}

The ultimate goal is to solve the $\ell_0$-constrained \textbf{Problem 1}~\eqref{eq:abstract_optimization_problem}, which seeks the best-fitting model with $n^{\mathcal{A}}_\phi$ non-zero coefficients.
However, solving the relaxed version introduced in \textbf{Problem 2}~\eqref{eq:l1_constrained_problem_abstract} is often preferred because of its proven ability to promote sparsity effectively \cite{flaschel_unsupervisedDiscoveryEUCLID_2021}. For instance, the $\ell_1$-norm was proposed in \cite{flaschel_brainEUCLID_2023} for efficiently handling the model discovery problem. It is worth noting that alternative $\ell_p$-penalties with $0<p<1$ can further approximate the $\ell_0$-penalty at the cost of non-convexity \cite{flaschel_unsupervisedDiscoveryEUCLID_2021,mcculloch_LPRegularizationModelDiscovery_2024,chartrand_reconstructionSparseSignals_2007,foucart_sparsestSolutionsLinearSystems_2009}.
In any case, these approaches require formulating \textbf{Problem 2}~\eqref{eq:l1_constrained_problem_abstract} as a linear regression problem to leverage efficient solvers \cite{hastie_statisticalLearning_2009}.

To formulate the regression problem, the pressure $p^{(k,d)}$ for each stress-deformation pair $d$ of deformation mode $k$ must be related to the unknown model parameters. Since $p^{(k,d)}$ is a Lagrange multiplier determined by the boundary conditions, it is also related to the \ac{SEF}, and the pressure becomes a function of $\vecc$ and $\vecw$:
\begin{equation}\label{eq:ansatz_pressure}
    p^{(k,d)} = \sum_{j=1}^{n_{\phi}} c_{j} \, p_{j}(\hat{\tenF}^{(k,d)}; \vecw_{j}),
\end{equation}
where $p_{j}$ is pressure contribution of the $j$-th basis function.
Substituting Eq.~\eqref{eq:ansatz_pressure} into Eq.~\eqref{eq:pk_tensor_model} yields:
\begin{equation} \label{eq:stress_rearranged}
    \tenP^{(k,d)} = \sum_{j=1}^{n_{\phi}} c_{j} \left[ \frac{\partial \phi_{j}(\hat{\tenF}^{(k,d)}; \vecw_{j})}{\partial \tenF} - p_{j}(\hat{\tenF}^{(k,d)}; \vecw_{j}) \,\hat{\tenF}^{(k,d)\minusTranspose} \right].
\end{equation}
The term in brackets is the effective stress contribution of the $j$-th basis function. 
The regression problem is formed by filtering the $n_{\mathrm{obs}}^{(k)}$ measurable stress components from the $n_{d}^{(k)}$ stress tensors $\tenP^{(k,d)}$ and assemble them into the unweighted subsystem for deformation mode $k$:
\begin{equation}\label{eq:pk_model_subsystem}
    {\vecP}^{(k)} = {\hat{\tenPsi}}^{(k)}(\hat{\mathcal{F}}^{(k)}; \vecw)  \vecc.
\end{equation}
Here, ${\vecP}^{(k)} \in \mathbb{R}^{n_{\mathrm{obs}}^{(k)}}$ comprises all predicted stresses for deformation mode $k$. Furthermore, ${\hat{\tenPsi}}^{(k)} \in \mathbb{R}^{n_{\mathrm{obs}}^{(k)} \times n_\phi}$ denotes the design matrix.
Each column of ${\hat{\tenPsi}}^{(k)}$ represents the stress contribution of a single basis function $\phi_{j}$ to the complete set of measurable stress components in a specific deformation mode, with the corresponding pressure effect already embedded within it.

To incorporate the weighting from the objective function, each unweighted subsystem is transformed into its final weighted form by pre-multiplying the weighting matrix $\hat{\mathbb{W}}^{(k)}$:
\begin{equation}\label{eq:weighted_subsystem}
    \hat{\vecP}^{(k)}_{\mathbb{W}} = \hat{\mathbb{W}}^{(k)} {\hat{\vecP}}^{(k)}, \quad \hat{\tenPsi}^{(k)}_{\mathbb{W}}(\hat{\mathcal{F}}^{(k)}; \vecw) = \hat{\mathbb{W}}^{(k)} {\hat{\tenPsi}}^{(k)}(\hat{\mathcal{F}}^{(k)}; \vecw).
\end{equation}
However, in what follows, we drop the $\mathbb{W}$ in the subscript of the weighted quantities for simplicity, but continue to refer to them unless otherwise stated.

The global design matrix $\hat{\tenPsi}(\vecw)$ and data vector $\hat{\vecP}$ are assembled by vertically stacking the blocks from each deformation mode, respectively, as follows:
\begin{equation}\label{eq:design_matrix_assembly}
    \hat{\tenPsi}(\vecw) = 
    \begin{bmatrix}
        \hat{\tenPsi}^{(1)}(\hat{\mathcal{F}}^{(1)};\vecw) \\
        \hat{\tenPsi}^{(2)}(\hat{\mathcal{F}}^{(2)};\vecw) \\
        \vdots \\
        \hat{\tenPsi}^{(n_{k})}(\hat{\mathcal{F}}^{(n_{k})};\vecw)
    \end{bmatrix},
    \quad
    \hat{\vecP} = 
    \begin{bmatrix}
        \hat{\vecP}^{(1)} \\
        \hat{\vecP}^{(2)} \\
        \vdots \\
        \hat{\vecP}^{(n_{k})}
    \end{bmatrix}.
\end{equation}
The design matrix $\hat{\tenPsi}(\vecw) \in \mathbb{R}^{n_{\mathrm{obs}} \times n_\phi}$, with $n_{\mathrm{obs}}$ as the total number of stress observations from the $n_{k}$ deformation modes, is a function of the non-linear parameters $\vecw$. 

Finally, we linearize and standardize the system. Linearization enables the efficient solution of the optimization problem, while standardization ensures equitable penalization in $\ell_1$-based methods and improves the numerical stability of $\ell_0$-based algorithms.
The linearization is performed by fixing the non-linear parameters to a set of pre-defined, constant values, $\vecw = \bar{\vecw}$. More details on how to select $\bar{\vecw}$ are provided in Section~\ref{subsec:anisotropic_analysis_final}. 
Consequently, the design matrix becomes a fully determined constant matrix, denoted as 
\begin{equation}\label{eq:linearized_design_matrix}
    \hat{\tenPsi}^{\bar{\vecw}} \equiv \hat{\tenPsi}(\vecw = \bar{\vecw}).
\end{equation} 

For the standardization, each column of $\hat{\tenPsi}^{\bar{\vecw}}$ is scaled to have a zero mean and unit variance. First, the mean $\mu_j$ and standard deviation $\sigma_j$ of each column $j$ are computed. Then, each column of the standardized matrix, $\tilde{\hat{\tenPsi}}^{\bar{\vecw}}$, is calculated as:
\begin{equation}
    \tilde{\hat{\tenPsi}}^{\bar{\vecw}}_{:,j} = \frac{\hat{\tenPsi}^{\bar{\vecw}}_{:,j} - \mu_j \mathbf{1}}{\sigma_j},
\end{equation}
where $\mathbf{1}$ is a vector of ones. Likewise, the data vector $\hat{\vecP}$ is centered by subtracting its mean value, $\mu_{\hat{P}}$:
\begin{equation}
    \tilde{\hat{\vecP}} = \hat{\vecP} - \mu_{\hat{P}} \mathbf{1}.
\end{equation}
For the sake of simplicity, in what follows, we drop the tilde from the standardized quantities, but we keep referring to them unless otherwise stated. The coefficients solved for in this standardized system will be denoted as $\tilde{\vecc}^{*}$ to distinguish them from the final, physical coefficients $\vecc^*$.

The sparse regression algorithms are then applied to this fully standardized system, solving the regression problem between $\hat{\vecP}$ and $\hat{\tenPsi}^{\bar{\vecw}}$ to find a vector of scaled optimal coefficients, $\tilde{\vecc}^*$. To recover the physically meaningful optimal coefficients, $\vecc^*$, for the unstandardized problem, a back-transformation is necessary. The physical coefficients are retrieved by:
\begin{equation}\label{eq:back_transformation_scaled_coefficients}
    c_j^* = \frac{\tilde{c}_j^*}{\sigma_j}.
\end{equation}

These two steps transform the non-linear optimization \textbf{Problem 1}~\eqref{eq:abstract_optimization_problem} into the following problem: 

\noindent \textbf{Problem 3: non-smooth $\ell_0$-constrained sparse linear regression problem}
\begin{equation}\label{eq:sparse_l0_linear_regression_problem}
    \tilde{\vecc}^*_{n^{\mathcal{A}}_\phi} = \arg\min_{\tilde{\vecc}} \left\| \hat{\vecP} - \hat{\tenPsi}^{\bar{\vecw}}  \tilde{\vecc} \right\|^2_2 \quad \text{subject to} \quad \|\tilde{\vecc}\|_0 \le n^{\mathcal{A}}_\phi,
\end{equation}
or, equivalently, the non-linear optimization \textbf{Problem 2}~\eqref{eq:l1_constrained_problem_abstract} into the following problem: 

\noindent \textbf{Problem 4: relaxed $\ell_1$-constrained sparse linear regression problem}
\begin{equation}\label{eq:sparse_l1_linear_regression_problem}
    \tilde{\vecc}^*_\tau = \arg\min_{\tilde{\vecc}} \left\| \hat{\vecP} - \hat{\tenPsi}^{\bar{\vecw}}  \tilde{\vecc} \right\|^2_2 \quad \text{subject to} \quad \|\tilde{\vecc}\|_1 \le \tau.
\end{equation}

Both optimization problems in Eqs.~\eqref{eq:sparse_l0_linear_regression_problem} and \eqref{eq:sparse_l1_linear_regression_problem} are the fundamental problems that the sparse regression algorithms we introduce in Section~\ref{subsec:sparse_regression_algo} are designed to approximate.
Notice that they do not yield a single coefficient vector directly.  
Instead, each produces a {path} of candidate solutions. For instance, \textbf{Problem~3} generates solutions parameterized by the sparsity level $n^{\mathcal{A}}_\phi$, while \textbf{Problem~4} generates solutions parameterized by the $\ell_1$-budget $\tau$. 
Recall that in Eqs.~\eqref{eq:sparse_l0_linear_regression_problem} and \eqref{eq:sparse_l1_linear_regression_problem}, $\hat{\vecP}$ and $\hat{\tenPsi}^{\bar{\vecw}}$ are the weighted quantities defined in Eq.~\eqref{eq:weighted_subsystem}.
The final scaled coefficient vector, $\tilde{\vecc}^*$, is then selected from this path using a model selection criterion; see Section~\ref{subsec:model_selection_criteria}. Finally, the selected scaled coefficient is back-transformed to the physical coefficient $\vecc^*$ according to Eq.~\eqref{eq:back_transformation_scaled_coefficients}.

\subsection{Sparse Regression Algorithms}
\label{subsec:sparse_regression_algo}

In this paper, we consider three well-known sparse regression algorithms: 

\paragraph{1. \ac{LASSO} (least absolute shrinkage and selection operator)}
The \ac{LASSO} \cite{tibshirani_regressionShrinkageSelectionLasso_1996} solves \textbf{Problem 4}~\eqref{eq:sparse_l1_linear_regression_problem} by addressing its equivalent penalized form:
\begin{equation}\label{eq:lasso_penalized_form}
    \tilde{\vecc}_{\lambda_L} = \arg\min_{\tilde{\vecc}} \left\{ \left\| \hat{\vecP} - \hat{\tenPsi}^{\bar{\vecw}}  \tilde{\vecc} \right\|^2_2 + \lambda_L \|\tilde{\vecc}\|_1 \right\}.
\end{equation}
Due to the convexity of the problem, {a one-to-one correspondence exists between the constraint budget $\tau$ in Eq.~\eqref{eq:sparse_l1_linear_regression_problem} 
and the regularization parameter $\lambda_L > 0$ in Eq.~\eqref{eq:lasso_penalized_form}. 
For $\lambda_L = 0$, the penalty vanishes and the problem reduces to the \ac{OLS} formulation.
Therefore, the constrained and penalized forms are equivalent under convex duality only for strictly positive $\lambda_L$ \cite{tibshirani_regressionShrinkageSelectionLasso_1996,osborne_lasso_2000}.}

{
In practice, the $\ell_1$-penalized problem~\eqref{eq:lasso_penalized_form} is solved using a convex optimization scheme based on coordinate descent \cite{friedman_pathwiseCoordinateOptimization_2007}. 
This iterative procedure updates one coefficient at a time while keeping the others fixed, efficiently handling the non-differentiability of the $\ell_1$-norm. 
The regularization parameter $\lambda_L$ controls the degree of sparsity: as $\lambda_L$ decreases, more coefficients become active, tracing the continuous LASSO path.} 
LASSO therefore generates a family of solutions $\tilde{\vecc}_{\lambda_L}$ along a decreasing sequence of $\lambda_L$ values. The final coefficient vector $\tilde{\vecc}^*$ are taken as $\tilde{\vecc}_{\lambda_L^\star}$ at the $\lambda_L^\star$ chosen by a model selection criterion.

\paragraph{2. \ac{LARS} (least angle regression)}
The \ac{LARS} algorithm \cite{efron_leastAngleRegression_2004} is a forward, stepwise procedure that traces the full path of solutions to \textbf{Problem 4}~\eqref{eq:sparse_l1_linear_regression_problem}. The algorithm is constructed to satisfy the \ac{KKT} optimality conditions of the \ac{LASSO} problem at each step.
Therefore, \ac{LASSO} and \ac{LARS} solve the same $\ell_1$-constrained problem. 

The procedure begins with an all-zero coefficient vector $\tilde{\vecc}_0 = \vec{0}$, an initial residual $\vecr_0 = \hat{\vecP}$, and an empty active set of features $\mathcal{A}_0 = \emptyset$. 
At each iteration $t$, the algorithm proceeds as follows:
\begin{enumerate}
    \item \textbf{Identify most correlated feature:} At the beginning of each iteration $t$, the algorithm identifies the feature most correlated with the residual from the prior step, $\vecr_{t-1}$. This selection is expressed as:
    \begin{equation*}
        j_t = \arg\max_{j} |(\hat{\tenPsi}^{\bar{\vecw}}_{:,j})^{\top} \, \vecr_{t-1}|, \quad \mathcal{A}_t = \mathcal{A}_{t-1} \cup \{j_t\}.
    \end{equation*}
    Here, the $\arg\max$ operation finds the {index} $j_t$ that maximizes the objective function across all features $j$ in the library. The objective function, $|(\hat{\tenPsi}^{\bar{\vecw}}_{:,j})^{\top} \, \vecr_{t-1}|$, is the absolute value of the {dot product} between a candidate feature vector (the $j$-th column of the design matrix $\hat{\tenPsi}^{\bar{\vecw}}$) and the residual vector. This value quantifies the magnitude of the {linear correlation} between each feature and the portion of the data unexplained by the model so far. The feature index $j_t$ with the highest correlation is then incorporated into the model by updating the active set via the set union operation $\mathcal{A}_t = \mathcal{A}_{t-1} \cup \{j_t\}$.
    Notice that the residual update here belongs to a step that {tracks the \ac{LASSO} solution path}, not an \ac{OLS} refit on the current support. This shrinkage might induce a bias in the estimates compared to the \ac{OLS} fit. The benefit is variance reduction and improved stability, particularly when predictors are collinear or when the number of predictors is large relative to the number of samples.

    \item \textbf{Move along equiangular path:} Update the coefficient vector by moving from $\tilde{\vecc}_{t-1}$ in a direction that is equiangular to all features in the current active set $\mathcal{A}_t$. The algorithm proceeds along this path until a new, inactive feature has the same correlation with the residual as those in $\mathcal{A}_t$. This process yields the updated coefficient vector $\tilde{\vecc}_t$. This specific direction ensures that the KKT conditions for the LASSO problem remain valid.
    
    \item \textbf{Residual update:} Calculate the new residual corresponding to the updated coefficients:
    \begin{equation*}
        \vecr_{t} = \hat{\vecP} - \hat{\tenPsi}^{\bar{\vecw}}  \tilde{\vecc}_{t}.
    \end{equation*}
\end{enumerate}
As iterations proceed, LARS produces a sequence of coefficient vectors 
$\{\tilde{\vecc}_t\}_{t=0}^{T}$, where $T$ denotes the final iteration index (typically bounded above by $n_{\phi}$). 
These iterates trace the piecewise-linear path associated with the \ac{LASSO} problem \cite{hastie_statisticalLearning_2009}. 
The final scaled coefficient vector is taken as $\tilde{\vecc}^* = \tilde{\vecc}_{t^\star}$, where the optimal iterate $t^\star$ along the LARS path is determined by the model selection criterion.

{In short, \ac{LARS} provides a constructive, geometric algorithm that generates the entire set of solutions to \textbf{Problem~4}~\eqref{eq:sparse_l1_linear_regression_problem} \cite{hastie_statisticalLearning_2009}.
Note that the solutions of \eqref{eq:sparse_l1_linear_regression_problem} may differ for \ac{LASSO} and \ac{LARS}.
In the absence of collinearity among features, the so-called \ac{LARS}--\ac{LASSO} modification yields the same solution path as \ac{LASSO} \cite{hastie_statisticalLearning_2009,efron_leastAngleRegression_2004,tibshirani_lassoUniqueness_2013}.
However, when predictors are strongly correlated, as is often the case in model libraries for constitutive model discovery, the \ac{LARS}--\ac{LASSO} may again deviate from the exact \ac{LASSO} trajectory. Therefore, in this work, we stick to the original \ac{LARS} algorithm.
}

\paragraph{3. \ac{OMP} (orthogonal matching pursuit)}
\ac{OMP} \cite{pati_orthogonalMatchingPursuit_1993} is a greedy algorithm that, in contrast to \ac{LASSO} and \ac{LARS}, finds an approximate solution to the $\ell_0$-constrained \textbf{Problem 3}~\eqref{eq:sparse_l0_linear_regression_problem}. To make this problem tractable, \ac{OMP} bypasses the combinatorial explosion of checking all possible feature subsets by employing a forward-selection heuristic. Starting with an all-zero coefficient vector $\tilde{\vecc}_0 = \vec{0}$, a residual $\vec{r}_0 = \hat{\vecP}$, and an empty active set $\mathcal{A}_0 = \emptyset$, \ac{OMP} performs the following steps at each iteration $t$:
\begin{enumerate}
    \item \textbf{Greedy selection:} Identify the feature most correlated with the current residual and add it to the active set. This selection is performed according to the expression:
    \begin{equation*}
        j_t = \arg\max_{j \notin \mathcal{A}_{t-1}} \frac{|(\hat{\tenPsi}^{\bar{\vecw}}_{:,j})^{\top} \, \vec{r}_{t-1}|}{\|\hat{\tenPsi}^{\bar{\vecw}}_{:,j}\|_2}, \quad \mathcal{A}_t = \mathcal{A}_{t-1} \cup \{j_t\}.
    \end{equation*}
    In this step, the $\arg\max$ operation searches for the $j_t$ feature that maximizes the objective function, restricted to all features $j$ that are not already in the active set from the previous iteration, $\mathcal{A}_{t-1}$. The objective function itself represents the {normalized correlation} between each candidate feature vector and the current residual vector $\vec{r}_{t-1}$. 
    This term quantifies the linear correlation between that feature and the portion of the data currently unexplained by the model. To ensure the selection is unbiased by the varying magnitudes of the feature vectors, this correlation is normalized by the denominator, $\|\hat{\tenPsi}^{\bar{\vecw}}_{:,j}\|_2$, which is the {$\ell_2$-norm} of the feature vector. Once the optimal index $j_t$ is identified, the active set for the current iteration is updated via the set union operation $\mathcal{A}_t = \mathcal{A}_{t-1} \cup \{j_t\}$. Specifically, the numerator, $|(\hat{\tenPsi}^{\bar{\vecw}}_{:,j})^{\top} \, \vec{r}_{t-1}|$, is the absolute value of the {dot product} between the $j$-th feature vector (the $j$-th column of the design matrix $\hat{\tenPsi}^{\bar{\vecw}}$) and the residual. 
    
    \item \textbf{Coefficient estimation:} Solve an unpenalized \ac{OLS} problem restricted to the features in the active set:
    \begin{equation*}
        \tilde{\vecc}^*_{\mathcal{A}_t} = \arg\min_{\tilde{\vecc}_{\mathcal{A}_t}} \left\|\hat{\vecP} - \hat{\tenPsi}^{\bar{\vecw}}_{\mathcal{A}_t}  \tilde{\vecc}_{\mathcal{A}_t}\right\|_2^2,
    \end{equation*}
    where $\hat{\tenPsi}^{\bar{\vecw}}_{\mathcal{A}_t}$ is the sub-matrix of $\hat{\tenPsi}^{\bar{\vecw}}$ containing the columns indexed by $\mathcal{A}_t$. The full coefficient vector $\tilde{\vecc}_t$ is then constructed by using the values from $\tilde{\vecc}^*_{\mathcal{A}_t}$ for the active features and zero for all others.

    \item \textbf{Residual update:} Calculate the new residual for the next iteration using the latest coefficient estimates:
    \begin{equation*}
        \vec{r}_t = \hat{\vecP} - \hat{\tenPsi}^{\bar{\vecw}}_{\mathcal{A}_t}  \tilde{\vecc}^*_{\mathcal{A}_t}.
    \end{equation*}
\end{enumerate}
\ac{OMP} proceeds until no inactive feature has a non-zero correlation with the residual, i.e.,
\begin{equation}
    \max_{j \notin \mathcal{A}_t} \left|(\hat{\tenPsi}^{\bar{\vecw}}_{:,j})^{\top}\vec{r}_t\right| = 0,
\end{equation}
which is equivalent to the residual being orthogonal to all remaining columns of $\hat{\tenPsi}^{\bar{\vecw}}$, or until all features have been added. This yields a nested sequence of active sets $\mathcal{A}_0 \subset \mathcal{A}_1 \subset \cdots \subset \mathcal{A}_T$ and corresponding coefficient vectors $\{\tilde{\vecc}_t\}_{t=0}^{T}$ tracing the full greedy path, with $T \le \mathrm{rank}(\hat{\tenPsi}^{\bar{\vecw}})$ in practice.

Notice that, in contrast to \ac{LASSO} and \ac{LARS}, \ac{OMP} performs an {exact \ac{OLS}} fit on the current active set at each iteration, then updates the residual. 
Because \ac{OMP} recomputes coefficients by full \ac{OLS} on the active set $\mathcal A_t$ at every step, the estimates are unbiased conditional on the support. But the greedy selection rule could produce higher variance and instability in the selected model, especially if features are correlated, since \ac{OMP} never shrinks coefficients and always increases the support monotonically. 

The final scaled coefficient vector is taken as $\tilde{\vecc}^* = \tilde{\vecc}_{t^\star}$, where the optimal iteration index $t^\star \in \{0,\dots,T\}$ is determined by a model selection criterion.

Table~\ref{tab:sparse_algorithms} summarizes the main features of the sparse regression algorithms introduced above. 

\begin{table}[h!]
\renewcommand{\arraystretch}{1.3}
\centering
\begin{tabular}{@{}p{2.5cm} p{4.2cm} p{4.2cm} p{4.2cm}@{}}
\toprule
& \textbf{LASSO} & \textbf{LARS} & \textbf{OMP} \\
\midrule
\textbf{Problem solved} 
& Solves \textbf{Problem 4} (Eq.~\eqref{eq:sparse_l1_linear_regression_problem}) for a predefined set of $\lambda_L$ values. 
& Traces the full solution path of \textbf{Problem 4} (Eq.~\eqref{eq:sparse_l1_linear_regression_problem}) using a forward, stepwise approach.
& Approximates \textbf{Problem 3} (Eq.~\eqref{eq:sparse_l0_linear_regression_problem}) using greedy forward selection. \\
\midrule
\textbf{Active set selection}
& \textbf{Implicitly} through coefficient shrinkage; features can enter and leave the active set.
& \textbf{Explicitly}; a feature enters when its correlation with the residual matches the active set. 
& \textbf{Explicitly}; adds the feature with the highest normalized correlation to the residual at each step. \\
\midrule
\textbf{Coefficient estimation}
& All coefficients are updated in each step, with many exactly equal to zero.
& Moves coefficients of the active set along a defined equiangular path.
& Estimates non-zero coefficients for the active set via an OLS fit. \\
\bottomrule
\end{tabular}
\caption{\textbf{Comparison of sparse regression algorithms:} LASSO, LARS, and OMP.}
\label{tab:sparse_algorithms}
\end{table}

\subsection{Model Selection Criteria}
\label{subsec:model_selection_criteria}

The model selection criteria considered herein aim to balance model fidelity with parsimony by penalizing unnecessary complexity while retaining predictive accuracy. All criteria are expressed in terms of the \ac{RSS}, which measures the squared $\ell_2$-norm of the discrepancy between measured stresses and model predictions for the optimized scaled coefficients $\tilde{\vecc}^*$:
\begin{equation}
    \text{RSS} = \left\| \hat{\vecP} - \hat{\tenPsi}^{\bar{\vecw}}  \tilde{\vecc}^* \right\|^2_2.
\end{equation}

\paragraph{\acf{AIC}.}  
The \ac{AIC} \cite{akaike_AkaikeInformationCriterion_2011} is rooted in information theory and provides an estimate of the expected Kullback–Leibler divergence between the data probability distribution and the probability distribution implied by the fitted model. Under Gaussian, homoscedastic error assumptions, it is given by
\begin{equation}
    \text{AIC} = n_{\mathrm{obs}} \ln{\frac{\text{RSS}}{n_{\mathrm{obs}}}} + 2 n^{\mathcal{A}}_\phi.
    \label{eq:aic_definition}
\end{equation}
The first term rewards goodness of fit by decreasing with smaller \ac{RSS}, while the second penalizes excessive model complexity. \ac{AIC} is asymptotically efficient for predictive performance, often selecting slightly more complex models than those obtained by stricter criteria \cite{vrieze_differenceAICAndBIC_2012}.

\paragraph{\acf{BIC}.}  
The \ac{BIC} \cite{neath_BayesianInformationCriterion_2012} arises from a large-sample approximation to the logarithm of the marginal likelihood under regular priors. It is formulated as
\begin{equation}
    \text{BIC} = n_{\mathrm{obs}} \ln{\frac{\text{RSS}}{n_{\mathrm{obs}}}} + n^{\mathcal{A}}_\phi \ln{n_{\mathrm{obs}}}.
    \label{eq:bic_definition}
\end{equation}
The second penalty term increases with $\ln{n_{\mathrm{obs}}}$, resulting in a stronger preference for parsimonious models than \ac{AIC}, particularly for large datasets. Under standard regularity conditions and assuming that the true model is contained in the candidate set, \ac{BIC} is consistent, selecting the true model with probability one as $n_{\mathrm{obs}} \to \infty$ \cite{schwarz_estimatingDimensionOfModel_1978}. 

\paragraph{\Acf{CV}.}  
\ac{CV} \cite{bengio_varianceKFoldCV_2003} offers a non-parametric approach to estimating generalization error directly from the data, without relying on large-sample approximations or likelihood assumptions. In $K$-fold \ac{CV}, the dataset is randomly partitioned into $K$ disjoint folds. For each fold $i$, the model is trained on the remaining $K\!-\!1$ folds to obtain $\tilde{\vecc}^{(i)}$, and its predictive error $CV_{\textrm{err}}$ is computed on the validation set $\mathcal{D}_i^{\text{val}}$:
\begin{equation}\label{eq:cv_error_definition}
    CV_{\textrm{err}} = \frac{1}{K} \sum_{i=1}^{K} \left( \frac{1}{N_{\textrm{obs},i}^{\textrm{val}}} \sum_{j \in \mathcal{D}_i^{\text{val}}} \left\| \hat{\vecP}^{(j)} - (\hat{\tenPsi}^{\bar{\vecw}})^{(j)}  \tilde{\vecc}^{(i)} \right\|_2^2 \right),
\end{equation}
where $N_{\textrm{obs},i}^{\textrm{val}}$ represents the number of observations in the $i$-th validation fold. Finally, the model complexity (e.g., $\lambda_L$ in \ac{LASSO} or $n^{\mathcal{A}}_\phi$ in \ac{OMP}) that minimizes $CV_{\textrm{err}}$ is chosen. 
$K\!\in\![5,10]$ provides a practical bias–variance trade-off \cite{kohavi_CVModelSelection_1995}.
Unlike \ac{AIC} and \ac{BIC}, \ac{CV} does not explicitly penalize complexity; instead, it discourages overfitting because models that perform poorly on unseen folds will have larger \ac{CV} error. In linear Gaussian problems, leave-one-out \ac{CV} yields results closely related to \ac{AIC} \cite{arlot_surveyCVModelSelection_2010}. 

Although all three approaches measure a fit-complexity trade-off, they target different optimality notions: \ac{AIC} estimates the model with minimum expected predictive risk, \ac{BIC} approximates the most probable model under a Bayesian framework, and \ac{CV} empirically estimates out-of-sample performance without strong parametric assumptions \cite{akaike_AkaikeInformationCriterion_2011,schwarz_estimatingDimensionOfModel_1978,arlot_surveyCVModelSelection_2010}.  
Table~\ref{tab:model_selection_criteria} summarizes the main features of the three model selection criteria considered in this section.

\begin{table}[h!]
\renewcommand{\arraystretch}{1.3}
\centering
\begin{tabular}{@{}p{2.5cm} p{4.2cm} p{4.2cm} p{4.2cm}@{}}
\toprule
& \textbf{AIC} & \textbf{BIC} & \textbf{CV} \\
\midrule
\textbf{Penalty on complexity}
& $2 n^{\mathcal{A}}_\phi$, linear in number of active terms 
& $n^{\mathcal{A}}_\phi \ln{n_{\mathrm{obs}}}$, stronger than AIC for large $n_{\mathrm{obs}}$ because of the presence of $\ln{n_{\mathrm{obs}}}$ 
& Implicit, through poor generalization on unseen validation folds \\
\midrule
\textbf{Selection tendency}
& Often favors slightly more complex models with good predictive accuracy 
& Strong preference for parsimonious models; consistent as $n_{\mathrm{obs}} \to \infty$
& Selects the model with the lowest estimated prediction error \\
\midrule
\textbf{Optimality notion}
& Minimizes expected Kullback–Leibler divergence (predictive risk) 
& Maximizes posterior model probability under the Bayesian framework 
& Empirical minimization of out-of-sample prediction error \\
\bottomrule
\end{tabular}
\caption{\textbf{Comparison of model selection criteria:} \ac{AIC}, \ac{BIC}, and $K$-fold \ac{CV}.}
\label{tab:model_selection_criteria}
\end{table}

In the present automated model discovery framework, these selection criteria replace subjective Pareto-front inspection by an objective decision metric. For each candidate model along the regularization path, the \ac{AIC}, \ac{BIC}, and \ac{CV} are evaluated, and the models minimizing them are selected \cite{hastie_statisticalLearning_2009}.

\subsection{Final Parameter Refinement}

Once a sparse model structure (i.e., an active set of basis functions $\mathcal{A}$) is identified by solving either linearized \textbf{Problem 3} or \textbf{Problem 4}, a final refinement step is performed. The purpose of this step is to obtain more accurate parameter estimates by solving a least-squares problem restricted to the selected features. Thereby, we aim to remove the bias induced by the sparsity-promoting penalty used in the discovery stage and allow for the fine-tuning of any non-linear parameters \cite{hastie_statisticalLearning_2009,efron_leastAngleRegression_2004}. The refinement is performed on the weighted, but unstandardized, physical system to yield the final coefficients $\vecc^*$. The nature of this refinement problem depends on the composition of the active set $\mathcal{A}$.

If the active set $\mathcal{A}$ contains only basis functions that are independent of any tunable non-linear parameters (i.e., the vector $\vecw_{\mathcal{A}}$ is empty), the refinement simplifies to a linear least-squares problem. In this work, an $\ell_2$ (Ridge) regularization is employed to enhance numerical stability, yielding the following optimization problem:
\begin{equation}
    \vecc_{\mathcal{A}}^* = \arg\min_{\vecc_{\mathcal{A}}} \left\{ \left\| \hat{\vecP} - \hat{\tenPsi}_{\mathcal{A}} \vecc_{\mathcal{A}} \right\|_2^2 + \lambda_{R} \left\|\vecc_{\mathcal{A}}\right\|_2^2 \right\}.
    \label{eq:linear_model_refinement}
\end{equation}
Here, $\hat{\tenPsi}_{\mathcal{A}}$ contains only the columns from the unstandardized design matrix corresponding to the active set, and $\lambda_R > 0$ is a small regularization parameter.

Conversely, if one or more basis functions in $\mathcal{A}$ possess tunable non-linear parameters $\vecw_{\mathcal{A}}$, the refinement becomes a non-linear least-squares problem which is the simultaneous optimization of both the linear coefficients $\vecc_{\mathcal{A}}$ and the non-linear parameters $\vecw_{\mathcal{A}}$:
\begin{equation}
    \bigl\{\vecc_{\mathcal{A}}^*, \vecw_{\mathcal{A}}^*\bigr\} = \arg\min_{\vecc_{\mathcal{A}}, \vecw_{\mathcal{A}}} \left\{ \left\| \hat{\vecP} - \hat{\tenPsi}_{\mathcal{A}}(\vecw_{\mathcal{A}}) \vecc_{\mathcal{A}} \right\|_2^2 + \lambda_{R} \left\|\vecc_{\mathcal{A}}\right\|_2^2 \right\}.
    \label{eq:nonlinear_model_refinement}
\end{equation}
This formulation allows for a more accurate fit by adjusting the internal non-linear parameters of the selected basis functions.

Following the refinement, a {hard thresholding} step is applied to promote additional sparsity and eliminate negligible contributions. Specifically, any coefficient with a magnitude smaller than $10^{-6}$ is set to zero, and the corresponding basis function is removed from $\mathcal{A}$. This numerical cutoff, which is well below the expected scale of the fitted coefficients, ensures that the final model retains only terms with a relevant effect on the predicted response. Hard thresholding after refinement is well-established in high-dimensional regression as a means to enhance interpretability while maintaining predictive accuracy \cite{blumensath_iterativeHardThresholding_2009, belloni_leastSquaresAfterModelSelection_2013}.

% \newpage
\section{Constitutive Model Libraries}
\label{sec:model_libraries}

The choice of basis functions $\{\phi_{j}\}_{j=1}^{n_\phi}$ contained in model libraries of type \eqref{eq:w_series_expansion} is crucial for the discovery problem. Herein, we introduce the model libraries for isotropic and anisotropic hyperelastic materials, which we later use in our numerical tests. We restrict our analysis to incompressible hyperelastic materials.

\subsection{Isotropic Model Library}
\label{subsec:isotropic_library}

In the isotropic scenario, we consider an isochoric model library $\overline{W}$ composed of generalized Mooney-Rivlin and Ogden model features. The abstract summation with a single index $j$ in Eq.~\eqref{eq:w_series_expansion} is now explicitly defined by separating the library into these two families of basis functions:
\begin{equation}\label{eq:model_library_iso}
    \overline{W}(\tenF;\vecc) = \sum_{\substack{j,k \ge 0 \\ 1 \le j+k \le n_{\phi}^{\textrm{MR}}}} c_{(j,k)} \phi^{\textrm{MR}}_{(j,k)}(\tenF) + \sum_{l=1}^{n_{\phi}^{\textrm{Ogden}}} c_{l} \phi^{\textrm{Ogden}}_{l}(\tenF),
\end{equation}
where $n_{\phi}^{\textrm{MR}}$ is the maximum order of the polynomial for the Mooney-Rivlin part (i.e., the maximum value for $j+k$). The first sum runs over all combinations of non-negative integers $(j,k) \in \mathbb{Z}_{\geq 0} $ satisfying the condition $1 \le j+k \le n_{\phi}^{\textrm{MR}}$. 
The basis functions are defined using the principal invariants of the right Cauchy-Green tensor, $\tenC = \tenF\transpose\tenF$, and the principal stretches, $\lambda_i$. For an incompressible material, the principal invariants are given by:
\begin{equation}\label{eq:principal_invariants}
\begin{aligned}
    I_1 &= \textrm{tr}(\tenC) \\
    I_2 &= \frac{1}{2}[(\textrm{tr}(\tenC))^2 - \textrm{tr}(\tenC^2)].
\end{aligned}
\end{equation}
The explicit form of the basis functions is:
\begin{equation}
\begin{aligned}
    \phi^{\textrm{MR}}_{(j,k)}(I_1, I_2) &= (I_1-3)^{j} (I_2-3)^{k}, \\
    \phi^{\textrm{Ogden}}_{(l)}(\lambda_1, \lambda_2, \lambda_3) &= \lambda_1^{\alpha^{(l)}} + \lambda_2^{\alpha^{(l)}} + \lambda_3^{\alpha^{(l)}} - 3.
\end{aligned}
\end{equation}
Herein, we consider fixed exponents
$\alpha^{(l)}\in\left[\alpha^{(1)}, ..., \alpha^{(n_{\phi}^{\textrm{Ogden}})} \right] \subset \mathbb{R}$.
The vector of free coefficients $\vecc$ contains all $c_{(j,k)}$ from the Mooney-Rivlin expansion up to the order $n_{\phi}^{\textrm{MR}}$ and all $n_{\phi}^{\textrm{Ogden}}$ coefficients $c_{l}$ for the Ogden part:
\begin{equation}\label{eq:example_structure_isotropic_parameters}
    \vecc = \left[c_{(1,0)}, c_{(0,1)}, \ldots, c_{(0,n_{\phi}^{\textrm{MR}})}, \ldots\ldots, c_{1}, \ldots, c_{n_{\phi}^{\textrm{Ogden}}}\right]\transpose.
\end{equation}
The model parameters are subject to the consistency conditions with linear elasticity \cite{holzapfel_nonlinearSolidMechanics_2000}:
\begin{equation}\label{eq:consistency_conditions}
\begin{aligned}
    \mu_0^{\textrm{MR}} &= 2\left( \frac{\partial {W}}{\partial I_1} + \frac{\partial {W}}{\partial I_2} \right) = 2(c_{(1,0)} + c_{(0,1)}) > 0, \\[1.0ex]
    \mu_0^{\textrm{Ogden}} &= \frac{1}{2} \sum_{l=1}^{n_{\phi}^{\textrm{Ogden}}} c_{l}\,\alpha^{(l)}(\alpha^{(l)}-1) > 0,
\end{aligned}
\end{equation}
where $c_{(1,0)}$ and $c_{(0,1)}$ are directly the coefficients of the basis functions $\phi^{\textrm{MR}}_{(1,0)}=I_1-3$ and $\phi^{\textrm{MR}}_{(0,1)}=I_2-3$, respectively, from the Mooney–Rivlin expansion in Eq.~\eqref{eq:model_library_iso}.
The total initial shear modulus for the combined model is $\mu_0 = \mu_0^{\textrm{MR}} + \mu_0^{\textrm{Ogden}}$, which must be positive to ensure the physical consistency of the constitutive material model  \cite{holzapfel_nonlinearSolidMechanics_2000}.

\subsection{Anisotropic Model Library}
\label{subsec:anisotropic_library}

In this work, we focus on the special case of orthotropy, which is relevant for biological tissues \cite{martonová_modelDiscoveryCardiacTissue_2025,holzapfel_biomechanicalPropertiesBiologicalTissues_2025}.
Orthotropy (as well as other classes of anisotropy) can be modeled using the concept of structural tensors $\tenM_i$ that are considered as an additional input to $\overline{W}$. Thus, we consider a \ac{SEF} of the type $\overline{W}(\tenF, \{\tenM_i\}_{i=1}^{n_M})$ for which it must hold:
\begin{equation}
    \overline{W}\left(\tenF, \{\tenM_i\}_{i=1}^{n_M}\right) = \overline{W} \left(\tenF  \tenQ\transpose,\,   \{\tenQ \tenM_i \tenQ\transpose\}_{i=1}^{n_M} \right).
\end{equation}
Here, $\{\tenM_i\}_{i=1}^{n_M}$ denotes a set of structural tensors that describe the class of material symmetry, and $\tenQ \in \mathcal{G}$ is any symmetry operation of the material symmetry group. Structural tensors, such as $\tenM_i = \veca_i \otimes \veca_i$ for a family of preferred directions $\veca_i$, encode the intrinsic material symmetry. Their inclusion in the \ac{SEF} allows for a unified and frame-indifferent representation of anisotropic effects, as all physical quantities can then be constructed from invariants of $\tenF$ and $\tenM_i$. In particular, this approach guarantees that the constitutive response respects the underlying material symmetry, as required for thermodynamic consistency and physical realism in anisotropic constitutive models \cite{kalina_NNMeetsAnisotropicHyperelasticity_2025}.

In the special case of human cardiac tissue, as modeled in~\cite{martonová_modelDiscoveryCardiacTissue_2025}, three mutually orthogonal preferred directions are defined in the reference configuration. 
These directions are encoded via the structural tensors
\begin{equation}\label{eq:structure_tensors_ortho}
    \tenM_{\textrm{f}} = \vecf_0 \otimes \vecf_0, \quad
    \tenM_{\textrm{s}} = \vecs_0 \otimes \vecs_0, \quad
    \tenM_{\textrm{n}} = \vecn_0 \otimes \vecn_0,
\end{equation}
corresponding to the local fiber, sheet, and normal directions, respectively. This choice enables the explicit incorporation of the underlying microstructural organization characteristic of the human cardiac tissue. 

The arguments of $\overline{W}$ are then constructed from the invariants of the Cauchy-Green tensor $\tenC$ defined in Eq.~\eqref{eq:principal_invariants} and a set of scalar invariants formed by combining $\tenC$ with these structural tensors which is ${\mathcal{I}_{\textrm{ortho}}} = \{{I}_{\textrm{4f}}, {I}_{\textrm{4s}}, {I}_{\textrm{4n}}, {I}_{\textrm{8fs}}, {I}_{\textrm{8fn}}, {I}_{\textrm{8sn}}\}$. The explicit form of these anisotropic invariants is provided in Appendix~\ref{sec:ortho_model}.

Here, we adopt the {orthotropic} model library for $\overline{W}$ proposed in \cite{martonová_modelDiscoveryCardiacTissue_2025}. The library utilizes the eight kinematic arguments derived from the isotropic and anisotropic invariants: 
\begin{equation}
    {\mathcal{I}} = \{{I}_1, {I}_2, I_{\textrm{4f}}, I_{\textrm{4s}}, I_{\textrm{4n}}, I_{\textrm{8fs}}, I_{\textrm{8fn}}, I_{\textrm{8sn}}\}.
\end{equation}
For each of these eight arguments, four types of basis functions (linear, exponential-linear, quadratic, exponential-quadratic) are considered, leading to $8 \times 4 = 32$ potential terms in the \ac{SEF}:
\begin{equation}\label{eq:sef_32terms_explicit_form_W_inc_updated}
    \overline{W}(\tenF, \tenM_{\textrm{f}} , \tenM_{\textrm{s}} , \tenM_{\textrm{n}} ; \vecc, \vecw) = \overline{W}(\mathcal{I}; \vecc, \vecw) = \sum_{j=1}^{n_\phi = 32} c_{j} \phi_{j}(\mathcal{I}; \vecw_{j}).
\end{equation}
Compared to the original \ac{SEF} proposed in \cite{martonová_modelDiscoveryCardiacTissue_2025}, we do not consider the linear and exponential-linear terms based on the corrected fourth invariants $\bigl[ \operatorname{max}\{I_{\textrm{4f}},1\} - 1\bigr]$, $\bigl[ \operatorname{max}\{I_{\textrm{4s}},1\} - 1\bigr]$, and $\bigl[ \operatorname{max}\{I_{\textrm{4n}},1\} - 1\bigr]$ as well as the linear and exponential-linear terms based on the eighth invariants $I_{\textrm{8fs}}$, $I_{\textrm{8fn}}$, and $I_{\textrm{8sn}}$, since these terms can be different from zero in deformation-free state and thereby induce undesired residual stresses. For a better comparison with the results, we used the same numbering of material parameters as in \cite{martonová_modelDiscoveryCardiacTissue_2025}. Ultimately, the \ac{SEF} ansatz in Eq.~\eqref{eq:sef_32terms_explicit_form_W_inc_updated} has a total of $\num{30}$ material parameters.
The explicit form of the basis functions $\phi_{j}$ is given in Appendix~\ref{sec:ortho_model}. The reader can also refer to \cite{martonová_modelDiscoveryCardiacTissue_2025} for more details.

% \newpage
\subsection{Deformation Modes and Stress-Strain Relationships}
\label{subsec:deformation_modes}

The relationship between the constitutive model parameters and the data is established by evaluating the stress response for specific deformation modes. This section details explicit stress-strain expressions used to construct the design matrix $\hat{\tenPsi}(\vecw)$ in Eq.~\eqref{eq:design_matrix_assembly}. In the component-wise expressions that follow, $F_{ij}$ and $P_{ij}$ denote the components of the deformation gradient tensor $\tenF$ and the first Piola-Kirchhoff stress tensor $\tenP$, respectively, with respect to a Cartesian coordinate system.

\subsubsection{Isotropic Materials}
\label{subsec:iso_def_modes}
The following deformation modes are relevant to the isotropic case. They are used both to generate synthetic data for benchmarking the sparse regression algorithms introduced in Section~\ref{sec:Methodology} and to model the experimental data from Treloar \cite{treloar_vulcanisedRubberData_1944}, which is widely regarded as prototypical for hyperelastic incompressible materials. For these tests, a plane stress assumption is commonly invoked to determine the hydrostatic pressure.

\paragraph{\Acf{UT}}
A stretch $\lambda$ is applied along the $X_1$ axis. Due to incompressibility and isotropy, the deformation gradient is: 
\begin{equation}
    \tenF_{\textrm{UT}}(\lambda) = \operatorname{diag}(\lambda, \lambda^{-1/2}, \lambda^{-1/2}).
\end{equation}
The pressure $p$ is determined from the boundary condition $P_{33}=0$, yielding the measurable nominal stress component $P_{11}$:
\begin{equation}
    P_{11} = \frac{\partial \overline{W}}{\partial F_{11}} - \frac{F_{33}}{F_{11}}\frac{\partial \overline{W}}{\partial F_{33}}.
\end{equation}

% \paragraph{\Acf{SS}}
% A shear of amount $\gamma$ is applied in the $X_1$-$X_2$ plane. The deformation gradient is:
% \begin{equation}
%     \tenF_{\textrm{SS}}(\gamma) = 
%     \begin{pmatrix} 1 & \gamma & 0 \\ 0 & 1 & 0 \\ 0 & 0 & 1 \end{pmatrix}.
% \end{equation}
% The off-diagonal shear stress component $P_{12}$ is independent of the hydrostatic pressure and is given by 
% \begin{equation}
%     P_{12} = \partial \overline{W} / \partial F_{12}.
% \end{equation}
% The normal stress components depend on the pressure, which is determined by enforcing a plane stress condition, $P_{33}=0$. This yields $p = \partial \overline{W} / \partial F_{33}$, resulting in the measurable normal stress components:
% \begin{align}
%     P_{11} &= \frac{\partial \overline{W}}{\partial F_{11}} - \frac{\partial \overline{W}}{\partial F_{33}}, \\
%     P_{22} &= \frac{\partial \overline{W}}{\partial F_{22}} - \frac{\partial \overline{W}}{\partial F_{33}}.
% \end{align}

\paragraph{\Acf{PS}}
In \ac{PS}, a stretch $\lambda$ is applied along $X_1$ while the $X_3$ dimension is held constant, yielding 
\begin{equation}
    \tenF_{\textrm{PS}}(\lambda) = \operatorname{diag}(\lambda,1, \lambda^{-1}).
\end{equation}
The pressure is determined from the condition $P_{33}=0$, which gives the measurable stress:
\begin{equation}
    P_{11} = \frac{\partial \overline{W}}{\partial F_{11}} - \frac{F_{33}}{F_{11}}\frac{\partial \overline{W}}{\partial F_{33}}.
\end{equation}

% \paragraph{Biaxial tension (BT and EBT)}
% Independent stretches $\lambda_1$ and $\lambda_2$ are applied along the $X_1$ and $X_2$ axes, respectively. For the general case (BT), the deformation gradient is:
% \begin{equation}
%     \tenF_{\textrm{BT}}(\lambda_1, \lambda_2) = \operatorname{diag}(\lambda_1, \lambda_2, (\lambda_1\lambda_2)^{-1}).
% \end{equation}
% For the \ac{EBT} case, $\lambda_1 = \lambda_2 = \lambda$. The pressure is found from the plane stress condition $P_{33}=0$. The two measurable stress components are then:
% \begin{align}
%     P_{11} &= \frac{\partial \overline{W}}{\partial F_{11}} - \frac{F_{33}}{F_{11}}\frac{\partial \overline{W}}{\partial F_{33}}, \\
%     P_{22} &= \frac{\partial \overline{W}}{\partial F_{22}} - \frac{F_{33}}{F_{22}}\frac{\partial \overline{W}}{\partial F_{33}}.
% \end{align}

\paragraph{Biaxial tension (BT and EBT)}
Independent stretches $\lambda_1$ and $\lambda_2$ are applied along the $X_1$ and $X_2$ axes, respectively. For the general case (BT), the deformation gradient is:
\begin{equation}
    \tenF_{\textrm{BT}}(\lambda_1, \lambda_2) = \operatorname{diag}(\lambda_1, \lambda_2, (\lambda_1\lambda_2)^{-1}).
\end{equation}
The pressure is determined from the plane-stress condition $P_{33}=0$. The two measurable stress components are then:
\begin{align}
    P_{11} &= \frac{\partial \overline{W}}{\partial F_{11}} - \frac{F_{33}}{F_{11}}\frac{\partial \overline{W}}{\partial F_{33}}, \\
    P_{22} &= \frac{\partial \overline{W}}{\partial F_{22}} - \frac{F_{33}}{F_{22}}\frac{\partial \overline{W}}{\partial F_{33}}.
\end{align}
For the \ac{EBT} case, $\lambda_1 = \lambda_2 = \lambda$, and, in consequence, $P_{11} = P_{22}$.

\subsubsection{Anisotropic Materials}
\label{subsec:aniso_def_modes}
The following deformation modes are relevant for the experimental data for human cardiac tissue from \cite{martonová_modelDiscoveryCardiacTissue_2025}. The loading axes are aligned with the principal material directions ($\vecf_0, \vecs_0, \vecn_0$). Accordingly, the tensor components are expressed in the material coordinate system, with indices $(\mathrm{f}, \mathrm{s}, \mathrm{n})$ instead of $(1, 2, 3)$.

\paragraph{\Acf{BT}}
This deformation is kinematically identical to the \ac{BT} described for isotropic materials. For an orthotropic material considered in this work, the stretches are applied along two of the principal material axes (e.g., fiber and normal), such that $\lambda_1 \equiv \lambda_{\mathrm{f}}$ and $\lambda_3 \equiv \lambda_{\mathrm{n}}$. The deformation gradient is therefore:
\begin{equation}
    \tenF_{\textrm{BT}}^{\textrm{ani}}(\lambda_{\mathrm{f}}, \lambda_{\mathrm{n}}) = \operatorname{diag}(\lambda_{\mathrm{f}}, (\lambda_{\mathrm{f}}\lambda_{\mathrm{n}})^{-1}, \lambda_{\mathrm{n}}).
\end{equation}
The pressure is determined from the zero-stress condition in the unconstrained sheet direction, $P_{\mathrm{ss}}=0$, yielding the two measurable stress components, $P_{\mathrm{ff}}$ and $P_{\mathrm{nn}}$:
\begin{align}
    P_{\mathrm{ff}} &= \frac{\partial \overline{W}}{\partial F_{\mathrm{ff}}} - \frac{F_{\mathrm{ss}}}{F_{\mathrm{ff}}}\frac{\partial \overline{W}}{\partial F_{\mathrm{ss}}}, \\
    P_{\mathrm{nn}} &= \frac{\partial \overline{W}}{\partial F_{\mathrm{nn}}} - \frac{F_{\mathrm{ss}}}{F_{\mathrm{nn}}}\frac{\partial \overline{W}}{\partial F_{\mathrm{ss}}}.
\end{align}

\paragraph{Triaxial simple shear}
Six modes of simple shear are considered, where a shear of amount $\gamma$ is applied in one of the principal material planes \cite{martonová_modelDiscoveryCardiacTissue_2025}. 
The deformation gradients for shear in the fiber-sheet ($\mathrm{fs}$), fiber-normal ($\mathrm{fn}$), and sheet-normal ($\mathrm{sn}$) planes are given by:
\begin{equation}
\begin{aligned}
    \tenF_{\textrm{SS}_{\mathrm{fs}}}(\gamma_{\mathrm{fs}}) &= 
    \begin{pmatrix} 1 & \gamma_{\mathrm{fs}} & 0 \\ 0 & 1 & 0 \\ 0 & 0 & 1 \end{pmatrix}, &
    \tenF_{\textrm{SS}_{\mathrm{sf}}}(\gamma_{\mathrm{sf}}) &= 
    \begin{pmatrix} 1 & 0 & 0 \\ \gamma_{\mathrm{sf}} & 1 & 0 \\ 0 & 0 & 1 \end{pmatrix}, \\
    \tenF_{\textrm{SS}_{\mathrm{fn}}}(\gamma_{\mathrm{fn}}) &= 
    \begin{pmatrix} 1 & 0 & \gamma_{\mathrm{fn}} \\ 0 & 1 & 0 \\ 0 & 0 & 1 \end{pmatrix}, &
    \tenF_{\textrm{SS}_{\mathrm{nf}}}(\gamma_{\mathrm{nf}}) &= 
    \begin{pmatrix} 1 & 0 & 0 \\ 0 & 1 & 0 \\ \gamma_{\mathrm{nf}} & 0 & 1 \end{pmatrix}, \\
    \tenF_{\textrm{SS}_{\mathrm{sn}}}(\gamma_{\mathrm{sn}}) &= 
    \begin{pmatrix} 1 & 0 & 0 \\ 0 & 1 & \gamma_{\mathrm{sn}} \\ 0 & 0 & 1 \end{pmatrix}, &
    \tenF_{\textrm{SS}_{\mathrm{ns}}}(\gamma_{\mathrm{ns}}) &= 
    \begin{pmatrix} 1 & 0 & 0 \\ 0 & 1 & 0 \\ 0 & \gamma_{\mathrm{ns}} & 1 \end{pmatrix}.
\end{aligned}
\end{equation}
\noindent For any simple shear mode, the off-diagonal (shear) components of the stress tensor are independent of the hydrostatic pressure. The measurable shear stress is therefore given directly by:
\begin{equation}
    P_{ij} = \frac{\partial \overline{W}}{\partial F_{ij}} \quad \text{for } i \neq j.
\end{equation}

% \newpage
\section{Numerical Results}\label{sec:numerical_results}

This section summarizes the numerical results obtained by applying the framework introduced in Section~\ref{sec:Methodology} to isotropic and anisotropic hyperelastic problems.
In Section~\ref{subsec:synthetic_benchmarks}, we benchmark the nine model discovery algorithms by considering four isotropic hyperelastic problems of increasing complexity. We investigate the ability of the algorithms to discover known ground truth models (Mooney-Rivlin, Ogden, and mixed forms) from synthetically generated data under various noise levels. 
In Section~\ref{subsec:Treloar}, we investigate the capabilities of the proposed algorithms for the isotropic experimental dataset from Treloar \cite{treloar_vulcanisedRubberData_1944} of vulcanized rubber.
Finally, in Section~\ref{subsec:anisotropic_analysis_final}, we apply the nine algorithms to an anisotropic hyperelastic case. We discover constitutive models for human cardiac tissue using the same experimental data as in \cite{martonová_modelDiscoveryCardiacTissue_2025}. The performance of the discovered anisotropic models is analyzed and compared with a four-term model presented in \cite{martonová_modelDiscoveryCardiacTissue_2025}. 

The implementation of the discovery algorithms relies on the core functionalities of the scikit-learn library \cite{pedregosa_Scikitlearn_2011}, leveraging its efficient solvers for \ac{LASSO}, \ac{LARS}, and \ac{OMP}.
{Specifically, the LASSO problem is solved using a coordinate descent algorithm that efficiently handles the non-differentiable $\ell_1$-penalty \cite{friedman_pathwiseCoordinateOptimization_2007}.
The LARS and OMP paths are computed using the iterative schemes outlined in Section~\ref{subsec:sparse_regression_algo}.}
In all numerical examples considered, we impose non-negativity of the regression coefficients. This design choice ensures that each invariant-based basis function contributes additively to the strain-energy density, preventing cancellation between terms. We have observed that this is particularly critical when the model library is constructed from Mooney-Rivlin and Ogden terms due to collinearity, and it serves as a sufficient condition to fulfill the consistency condition in Eq.~\eqref{eq:consistency_conditions}. Additionally, non-negativity of the coefficients is a necessary condition for the anisotropic model library to remain physically consistent \cite{martonová_modelDiscoveryCardiacTissue_2025}. 
We evaluate the performance of the discovered models in each case using the coefficient of determination ($R^2$) and the \ac{RMSE}.
The source code developed for this study, along with the data required to reproduce our findings, is publicly available on Zenodo \cite{urrea_codeAutomatedModelDiscovery_2025}.

\subsection{Benchmarking - Isotropic Synthetic Data}
\label{subsec:synthetic_benchmarks}

Four isotropic hyperelastic models are analyzed: \textbf{i.} a two-term Ogden model (O2), \textbf{ii.} a standard two-term Mooney-Rivlin model (MR2), and \textbf{iii-iv.} two mixed Mooney-Rivlin/Ogden models (MR1O1, MR2O2). Synthetic data covering \ac{UT}, \ac{PS}, and \ac{EBT} modes are generated for principal stretches $\lambda \in [0.6, 5.0]$. For all three deformation modes, we assume that only the stress component $P_{11}$ is measured and choose $n_{\mathrm{obs}}^{(k)} = 60$. Gaussian noise, with standard deviations proportional to the true stress values (0\%, 5\%, and 10\% of the true stress), was added to the stress components to simulate experimental errors. For each synthetic test case, the library provided to the discovery algorithms contained only terms relevant to the ground truth model's family (e.g., only Mooney-Rivlin terms for MR2, only Ogden terms for O2, and a mix for the MR-Ogden models). 

The results, summarized in Table~\ref{tab:synth_performance_summary_full_kpa_time_rel}, confirm the high efficacy of the nine algorithms for the different benchmark problems. For the O2 model, in the absence of noise, all nine algorithms perfectly identified the correct two-term structure. This success was largely replicated under noisy conditions, where all information-criterion-based methods (\ac{LASSO}-\ac{AIC}/\ac{BIC}, \ac{LARS}-\ac{AIC}/\ac{BIC}, \ac{OMP}-\ac{AIC}/\ac{BIC}) found the ground truth form at both 5\% and 10\% noise levels. \ac{CV}-based algorithms also correctly discovered the two-term ground truth in nearly all cases, with the exception of \ac{LASSO}-\ac{CV} and \ac{LARS}-\ac{CV} at 10\% noise.

For the MR2 benchmark, \ac{LASSO}-\ac{CV}/\ac{AIC}/\ac{BIC} and \ac{OMP}-based algorithms are exact at 0\% noise, while \ac{LARS}-based methods struggled, selecting only one of the two terms. With the introduction of 5-10\% noise, all nine algorithms became highly accurate, consistently identifying the correct two-term model after the refit and thresholding procedure.
In the MR1O1 case, all nine algorithms perfectly identified the correct two-term structure.

The discovery of the four-term MR2O2 model represents the most demanding test. At 0\% and 5\% noise, all nine algorithms successfully recovered the ground truth. With 10\% noise, \ac{LASSO}-CV identified one additional spurious term, the Ogden term $\sum_k (\lambda_k^3 - 1)$. However, the discovered model remained highly accurate ($R^2_{\min}\approx 0.9786$).

\begin{figure}[htbp]
    \centering
    \begin{overpic}[width=0.98\textwidth,abs]{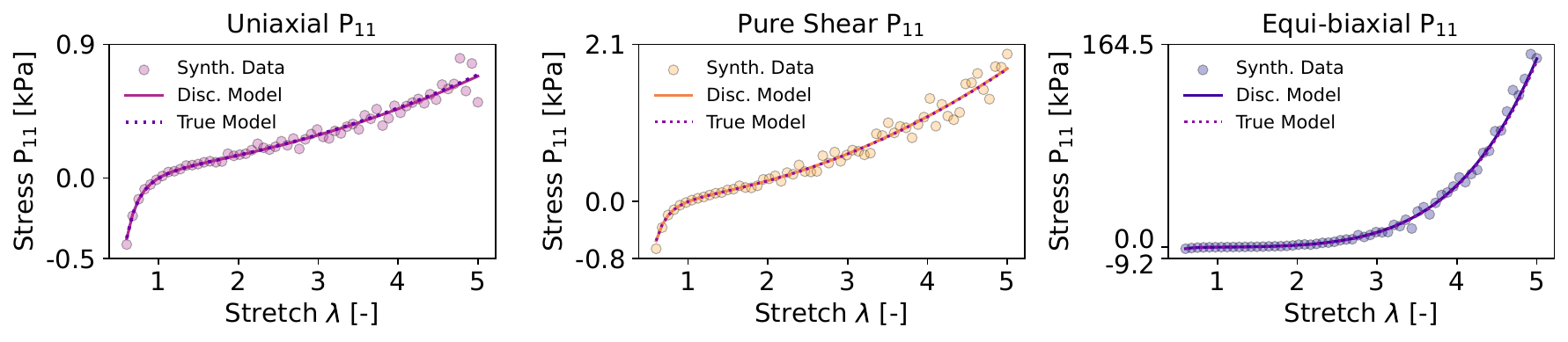}
        \put(1,110){\textbf{a. O2}}
    \end{overpic}
    \vspace{2em}
    
    \begin{overpic}[width=0.98\textwidth,abs]{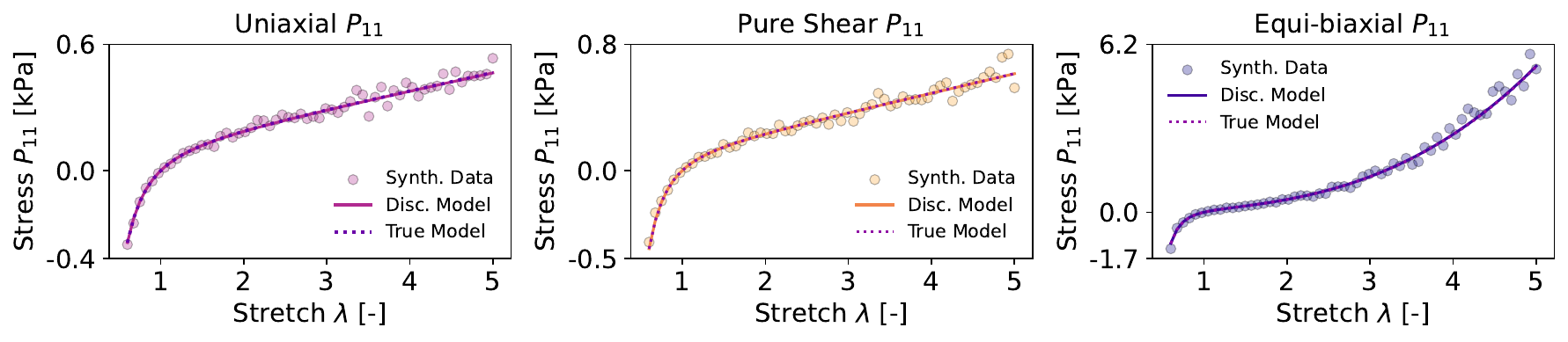}
        \put(1,110){\textbf{b. MR2}}
    \end{overpic}
    \vspace{2em}
    
    \begin{overpic}[width=0.98\textwidth,abs]{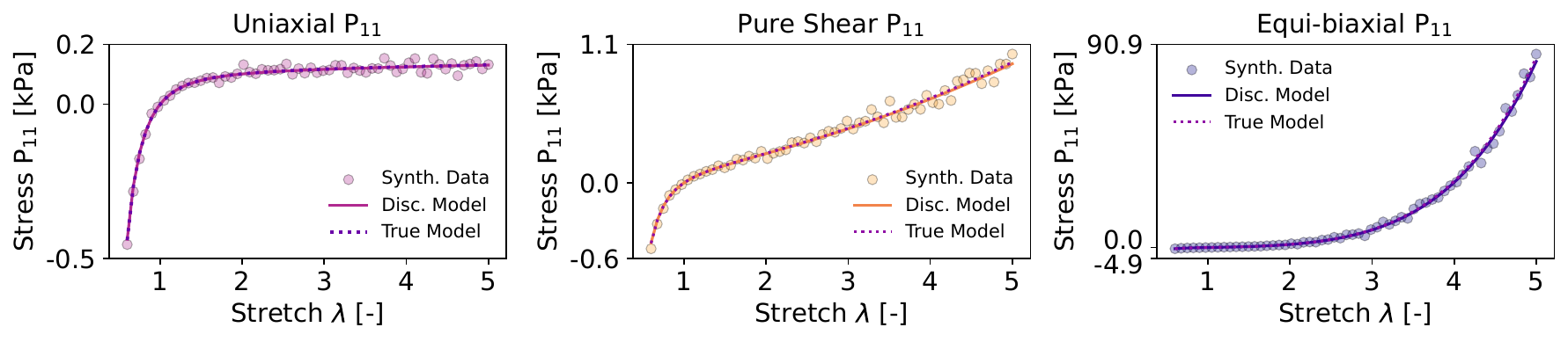}
        \put(1,110){\textbf{c. MR1O1}}
    \end{overpic}
    \vspace{2em}

    \begin{overpic}[width=0.98\textwidth,abs]{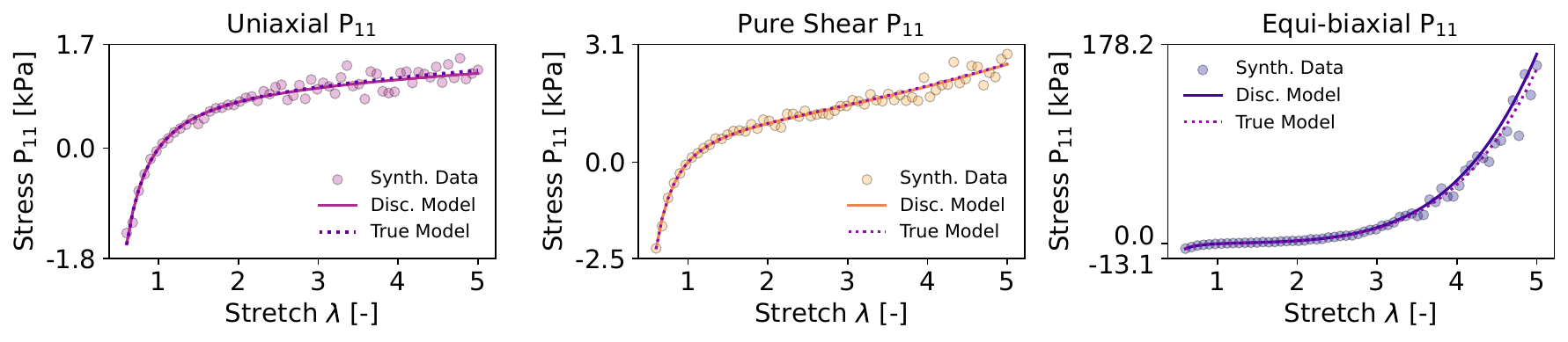}
        \put(1,110){\textbf{d. MR2O2}}
    \end{overpic}
    
    \caption{\textbf{Isotropic Synthetic Data:} Predicted stress-stretch responses for the best discovered models under 10\% relative noise, corresponding to the data in Table~\ref{tab:synth_best_model_forms_kpa_nrmse_rel}: 
    \textbf{(a)} O2 ground truth, discovered by \ac{LASSO}-\ac{AIC} (2 terms).
    \textbf{(b)} MR2 ground truth, discovered by \ac{LASSO}-\ac{CV} (2 terms).
    \textbf{(c)} MR1O1 ground truth, discovered by \ac{LASSO}-\ac{BIC} (2 terms).
    \textbf{(d)} MR2O2 ground truth, discovered by \ac{LASSO}-\ac{AIC} (4 terms).
    Ground truth responses are shown for comparison.}
    \label{fig:best_rediscovered_models_10pct_noise}
\end{figure}

These results show that the nine algorithms are robust for discovering material constitutive models. That is, for the majority of the benchmarks, the ground truth model is discovered with nearly identical performance. For the sake of brevity, a single representative example of the explicit form of the \ac{SEF} discovered with the highest performance metrics for each benchmark case is presented in Table~\ref{tab:synth_best_model_forms_kpa_nrmse_rel}. The performance of the best-identified models for each case with 10\% noise is visualized in Figure~\ref{fig:best_rediscovered_models_10pct_noise}.

\newpage
{\footnotesize
\begin{longtable}{@{}lllcccccccc@{}}
\caption{\textbf{Isotropic Synthetic Data:} Performance summary of discovered models . $R^2$ values are the coefficients of determination of the $P_{11}$ stress component for the \acf{UT}, \acf{PS}, and \acf{EBT} deformation modes. AvgNRMSE is the averaged normalized root mean squared error across the three deformation modes. Both the $R^2$ values and AvgNRMSE are calculated against the clean data. Time is the initial sparse identification duration in seconds.}
\label{tab:synth_performance_summary_full_kpa_time_rel} \\
\toprule
Scenario & Algorithm & \begin{tabular}[c]{@{}l@{}}Selection\\ criteria\end{tabular} & \begin{tabular}[c]{@{}c@{}}Ground truth\\ discovered?\end{tabular} & $n^{\mathcal{A}}_{\phi}$ & $R^2_{\text{UT}}$ & $R^2_{\text{PS}}$ & $R^2_{\text{EBT}}$ & AvgNRMSE & \begin{tabular}[c]{@{}c@{}}Time\\ {[}s{]}\end{tabular} \\
\midrule
\endfirsthead
\toprule
Scenario & Algorithm & \begin{tabular}[c]{@{}l@{}}Selection\\ criteria\end{tabular} & \begin{tabular}[c]{@{}c@{}}Ground truth\\ discovered?\end{tabular} & $n^{\mathcal{A}}_{\phi}$ & $R^2_{\text{UT}}$ & $R^2_{\text{PS}}$ & $R^2_{\text{EBT}}$ & AvgNRMSE & \begin{tabular}[c]{@{}c@{}}Time\\ {[}s{]}\end{tabular} \\
\midrule
\endhead
\bottomrule
\endfoot
O2 & LASSO & CV & $\checkmark$ & 2 & 1.0000 & 1.0000 & 1.0000 & 0.0000 & 0.4717 \\
0\% noise & & AIC & $\checkmark$ & 2 & 1.0000 & 1.0000 & 1.0000 & 0.0000 & 0.0069 \\
& & BIC & $\checkmark$ & 2 & 1.0000 & 1.0000 & 1.0000 & 0.0000 & 0.0069 \\
& LARS & CV & $\checkmark$ & 2 & 1.0000 & 1.0000 & 1.0000 & 0.0000 & 0.0033 \\
& & AIC & $\checkmark$ & 2 & 1.0000 & 1.0000 & 1.0000 & 0.0000 & 0.0052 \\
& & BIC & $\checkmark$ & 2 & 1.0000 & 1.0000 & 1.0000 & 0.0000 & 0.0053 \\
& OMP & CV & $\checkmark$ & 2 & 1.0000 & 1.0000 & 1.0000 & 0.0000 & 0.0035 \\
& & AIC & $\checkmark$ & 2 & 1.0000 & 1.0000 & 1.0000 & 0.0000 & 0.0035 \\
& & BIC & $\checkmark$ & 2 & 1.0000 & 1.0000 & 1.0000 & 0.0000 & 0.0035 \\ \midrule
O2 & LASSO & CV & $\checkmark$ & 2 & 0.9998 & 1.0000 & 0.9999 & 0.0020 & 0.5114 \\
5\% noise & & AIC & $\checkmark$ & 2 & 0.9996 & 1.0000 & 0.9998 & 0.0027 & 0.0066 \\
& & BIC & $\checkmark$ & 2 & 0.9996 & 1.0000 & 0.9998 & 0.0027 & 0.0066 \\
& LARS & CV & $\checkmark$ & 2 & 0.9996 & 1.0000 & 0.9998 & 0.0027 & 0.0280 \\
& & AIC & $\checkmark$ & 2 & 0.9998 & 1.0000 & 0.9999 & 0.0020 & 0.0376 \\
& & BIC & $\checkmark$ & 2 & 0.9999 & 1.0000 & 0.9998 & 0.0024 & 0.0377 \\
& OMP & CV & $\checkmark$ & 2 & 0.9999 & 1.0000 & 0.9999 & 0.0018 & 0.0122 \\
& & AIC & $\checkmark$ & 2 & 0.9999 & 1.0000 & 0.9999 & 0.0018 & 0.0122 \\
& & BIC & $\checkmark$ & 2 & 0.9999 & 1.0000 & 0.9998 & 0.0024 & 0.0122 \\ \midrule
O2 & LASSO & CV & $\times$ & 3 & 1.0000 & 0.9999 & 0.9998 & 0.0027 & 0.4172 \\
10\% noise & & AIC & $\checkmark$ & 2 & 1.0000 & 0.9999 & 0.9999 & 0.0017 & 0.0065 \\
& & BIC & $\checkmark$ & 2 & 1.0000 & 0.9999 & 0.9999 & 0.0017 & 0.0065 \\
& LARS & CV & $\times$ & 3 & 0.9999 & 0.9997 & 0.9995 & 0.0041 & 0.0272 \\
& & AIC & $\checkmark$ & 2 & 1.0000 & 0.9999 & 0.9999 & 0.0017 & 0.0366 \\
& & BIC & $\checkmark$ & 2 & 1.0000 & 0.9999 & 0.9999 & 0.0017 & 0.0367 \\
& OMP & CV & $\checkmark$ & 2 & 1.0000 & 0.9999 & 0.9999 & 0.0017 & 0.0094 \\
& & AIC & $\checkmark$ & 2 & 1.0000 & 0.9999 & 0.9999 & 0.0017 & 0.0094 \\
& & BIC & $\checkmark$ & 2 & 1.0000 & 0.9999 & 0.9999 & 0.0017 & 0.0094 \\ \midrule
MR2 & LASSO & CV & $\checkmark$ & 2 & 1.0000 & 1.0000 & 1.0000 & 0.0000 & 0.4322 \\
0\% noise & & AIC & $\checkmark$ & 2 & 1.0000 & 1.0000 & 1.0000 & 0.0000 & 0.0058 \\
& & BIC & $\checkmark$ & 2 & 1.0000 & 1.0000 & 1.0000 & 0.0000 & 0.0058 \\
& LARS & CV & $\times$ & 1 & 0.9157 & 0.6670 & -0.6002 & 0.1656 & 0.0035 \\
& & AIC & $\times$ & 1 & 0.9157 & 0.6670 & -0.6002 & 0.1656 & 0.0053 \\
& & BIC & $\times$ & 1 & 0.9157 & 0.6670 & -0.6002 & 0.1656 & 0.0054 \\
& OMP & CV & $\checkmark$ & 2 & 1.0000 & 1.0000 & 1.0000 & 0.0000 & 0.0050 \\
& & AIC & $\checkmark$ & 2 & 1.0000 & 1.0000 & 1.0000 & 0.0000 & 0.0050 \\
& & BIC & $\checkmark$ & 2 & 1.0000 & 1.0000 & 1.0000 & 0.0000 & 0.0050 \\ \midrule
MR2 & LASSO & CV & $\checkmark$ & 2 & 1.0000 & 1.0000 & 0.9999 & 0.0014 & 0.5319 \\
5\% noise & & AIC & $\checkmark$ & 2 & 1.0000 & 1.0000 & 1.0000 & 0.0009 & 0.0063 \\
& & BIC & $\checkmark$ & 2 & 1.0000 & 1.0000 & 1.0000 & 0.0009 & 0.0063 \\
& LARS & CV & $\checkmark$ & 2 & 1.0000 & 1.0000 & 0.9998 & 0.0017 & 0.0207 \\
& & AIC & $\checkmark$ & 2 & 1.0000 & 1.0000 & 0.9998 & 0.0017 & 0.0282 \\
& & BIC & $\checkmark$ & 2 & 1.0000 & 1.0000 & 0.9998 & 0.0017 & 0.0283 \\
& OMP & CV & $\checkmark$ & 2 & 1.0000 & 1.0000 & 1.0000 & 0.0009 & 0.0064 \\
& & AIC & $\checkmark$ & 2 & 1.0000 & 1.0000 & 0.9998 & 0.0017 & 0.0064 \\
& & BIC & $\checkmark$ & 2 & 1.0000 & 1.0000 & 0.9998 & 0.0017 & 0.0064 \\ \midrule
MR2 & LASSO & CV & $\checkmark$ & 2 & 0.9999 & 1.0000 & 0.9999 & 0.0019 & 0.4203 \\
10\% noise & & AIC & $\checkmark$ & 2 & 0.9999 & 1.0000 & 0.9998 & 0.0020 & 0.0057 \\
& & BIC & $\checkmark$ & 2 & 0.9999 & 1.0000 & 0.9998 & 0.0020 & 0.0057 \\
& LARS & CV & $\checkmark$ & 2 & 0.9998 & 1.0000 & 0.9998 & 0.0027 & 0.0208 \\
& & AIC & $\checkmark$ & 2 & 0.9998 & 1.0000 & 0.9998 & 0.0027 & 0.0282 \\
& & BIC & $\checkmark$ & 2 & 0.9998 & 1.0000 & 0.9998 & 0.0027 & 0.0283 \\
& OMP & CV & $\checkmark$ & 2 & 0.9999 & 1.0000 & 0.9999 & 0.0019 & 0.0078 \\
& & AIC & $\checkmark$ & 2 & 0.9999 & 1.0000 & 0.9998 & 0.0020 & 0.0078 \\
& & BIC & $\checkmark$ & 2 & 0.9999 & 1.0000 & 0.9998 & 0.0020 & 0.0078 \\ \midrule
MR1O1 & LASSO & CV & $\checkmark$ & 2 & 1.0000 & 1.0000 & 1.0000 & 0.0000 & 0.4330 \\
0\% noise & & AIC & $\checkmark$ & 2 & 1.0000 & 1.0000 & 1.0000 & 0.0000 & 0.0060 \\
& & BIC & $\checkmark$ & 2 & 1.0000 & 1.0000 & 1.0000 & 0.0000 & 0.0060 \\
& LARS & CV & $\checkmark$ & 2 & 1.0000 & 1.0000 & 1.0000 & 0.0000 & 0.0034 \\
& & AIC & $\checkmark$ & 2 & 1.0000 & 1.0000 & 1.0000 & 0.0000 & 0.0052 \\
& & BIC & $\checkmark$ & 2 & 1.0000 & 1.0000 & 1.0000 & 0.0000 & 0.0053 \\
& OMP & CV & $\checkmark$ & 2 & 1.0000 & 1.0000 & 1.0000 & 0.0000 & 0.0052 \\
& & AIC & $\checkmark$ & 2 & 1.0000 & 1.0000 & 1.0000 & 0.0000 & 0.0052 \\
& & BIC & $\checkmark$ & 2 & 1.0000 & 1.0000 & 1.0000 & 0.0000 & 0.0052 \\ \midrule
MR1O1 & LASSO & CV & $\checkmark$ & 2 & 0.9993 & 0.9996 & 0.9950 & 0.0093 & 0.4468 \\
5\% noise & & AIC & $\checkmark$ & 2 & 1.0000 & 1.0000 & 1.0000 & 0.0010 & 0.0081 \\
& & BIC & $\checkmark$ & 2 & 1.0000 & 1.0000 & 1.0000 & 0.0010 & 0.0081 \\
& LARS & CV & $\checkmark$ & 2 & 1.0000 & 1.0000 & 1.0000 & 0.0010 & 0.0497 \\
& & AIC & $\checkmark$ & 2 & 0.9994 & 0.9998 & 0.9963 & 0.0080 & 0.0679 \\
& & BIC & $\checkmark$ & 2 & 0.9997 & 0.9999 & 0.9983 & 0.0054 & 0.0680 \\
& OMP & CV & $\checkmark$ & 2 & 0.9995 & 0.9998 & 0.9970 & 0.0073 & 0.0120 \\
& & AIC & $\checkmark$ & 2 & 0.9995 & 0.9998 & 0.9970 & 0.0073 & 0.0120 \\
& & BIC & $\checkmark$ & 2 & 0.9997 & 0.9999 & 0.9983 & 0.0054 & 0.0120 \\ \midrule
MR1O1 & LASSO & CV & $\checkmark$ & 2 & 0.9997 & 0.9997 & 0.9966 & 0.0075 & 0.4151 \\
10\% noise & & AIC & $\checkmark$ & 2 & 0.9996 & 0.9998 & 0.9973 & 0.0069 & 0.0066 \\
& & BIC & $\checkmark$ & 2 & 1.0000 & 1.0000 & 1.0000 & 0.0007 & 0.0066 \\
& LARS & CV & $\checkmark$ & 2 & 0.9997 & 0.9997 & 0.9966 & 0.0075 & 0.0500 \\
& & AIC & $\checkmark$ & 2 & 0.9997 & 0.9997 & 0.9966 & 0.0075 & 0.0669 \\
& & BIC & $\checkmark$ & 2 & 0.9996 & 0.9998 & 0.9973 & 0.0069 & 0.0670 \\
& OMP & CV & $\checkmark$ & 2 & 0.9998 & 0.9998 & 0.9975 & 0.0065 & 0.0098 \\
& & AIC & $\checkmark$ & 2 & 0.9997 & 0.9997 & 0.9966 & 0.0075 & 0.0098 \\
& & BIC & $\checkmark$ & 2 & 0.9998 & 0.9998 & 0.9975 & 0.0065 & 0.0098 \\ \midrule
MR2O2 & LASSO & CV & $\checkmark$ & 4 & 1.0000 & 0.9999 & 1.0000 & 0.0013 & 0.4287 \\
0\% noise & & AIC & $\checkmark$ & 4 & 1.0000 & 1.0000 & 1.0000 & 0.0000 & 0.0078 \\
& & BIC & $\checkmark$ & 4 & 1.0000 & 1.0000 & 1.0000 & 0.0000 & 0.0078 \\
& LARS & CV & $\checkmark$ & 4 & 1.0000 & 1.0000 & 1.0000 & 0.0000 & 0.0062 \\
& & AIC & $\checkmark$ & 4 & 1.0000 & 1.0000 & 1.0000 & 0.0000 & 0.0089 \\
& & BIC & $\checkmark$ & 4 & 1.0000 & 1.0000 & 1.0000 & 0.0000 & 0.0090 \\
& OMP & CV & $\checkmark$ & 4 & 1.0000 & 1.0000 & 1.0000 & 0.0000 & 0.0135 \\
& & AIC & $\checkmark$ & 4 & 1.0000 & 1.0000 & 1.0000 & 0.0000 & 0.0135 \\
& & BIC & $\checkmark$ & 4 & 1.0000 & 1.0000 & 1.0000 & 0.0000 & 0.0135 \\ \midrule
MR2O2 & LASSO & CV & $\checkmark$ & 4 & 0.9969 & 0.9996 & 0.9975 & 0.0096 & 0.4602 \\
5\% noise & & AIC & $\checkmark$ & 4 & 1.0000 & 0.9997 & 0.9984 & 0.0051 & 0.0088 \\
& & BIC & $\checkmark$ & 4 & 1.0000 & 0.9997 & 0.9994 & 0.0037 & 0.0088 \\
& LARS & CV & $\checkmark$ & 4 & 0.9998 & 0.9997 & 0.9973 & 0.0066 & 0.0520 \\
& & AIC & $\checkmark$ & 4 & 0.9998 & 0.9997 & 0.9973 & 0.0066 & 0.0700 \\
& & BIC & $\checkmark$ & 4 & 0.9999 & 0.9997 & 0.9994 & 0.0042 & 0.0701 \\
& OMP & CV & $\checkmark$ & 4 & 0.9998 & 0.9997 & 0.9972 & 0.0067 & 0.0182 \\
& & AIC & $\checkmark$ & 4 & 0.9998 & 0.9997 & 0.9972 & 0.0067 & 0.0182 \\
& & BIC & $\checkmark$ & 4 & 0.9998 & 0.9997 & 0.9972 & 0.0067 & 0.0182 \\ \midrule
MR2O2 & LASSO & CV & $\times$ & 5 & 0.9989 & 0.9997 & 0.9786 & 0.0162 & 0.4546 \\
10\% noise & & AIC & $\checkmark$ & 4 & 0.9990 & 0.9997 & 0.9991 & 0.0059 & 0.0108 \\
& & BIC & $\checkmark$ & 4 & 0.9990 & 0.9997 & 0.9991 & 0.0059 & 0.0108 \\
& LARS & CV & $\checkmark$ & 4 & 0.9988 & 0.9998 & 0.9812 & 0.0153 & 0.0527 \\
& & AIC & $\checkmark$ & 4 & 0.9988 & 0.9998 & 0.9812 & 0.0153 & 0.0712 \\
& & BIC & $\checkmark$ & 4 & 0.9988 & 0.9998 & 0.9812 & 0.0153 & 0.0713 \\
& OMP & CV & $\checkmark$ & 4 & 0.9989 & 0.9998 & 0.9870 & 0.0131 & 0.0194 \\
& & AIC & $\checkmark$ & 4 & 0.9988 & 0.9998 & 0.9845 & 0.0141 & 0.0194 \\
& & BIC & $\checkmark$ & 4 & 0.9989 & 0.9998 & 0.9870 & 0.0131 & 0.0194 \\
\end{longtable}
}

\begin{table}[htbp!]
\centering
\caption{\textbf{Isotropic Synthetic Data:} Ground truth \acp{SEF} and best discovered models.}
\label{tab:synth_best_model_forms_kpa_nrmse_rel}
\resizebox{0.97\textwidth}{!}{%
\small
\begin{tabular}{@{}l l cc@{}}
\toprule
\multicolumn{1}{@{}l}{\begin{tabular}[c]{@{}l@{}}Model / \\ Noise Level (Algorithm, $n^{\mathcal{A}}_{\phi}$)\end{tabular}} &
\multicolumn{1}{l}{Explicit SEF Form ($\overline{W}$)} &
$R^2_{\min}$ & AvgNRMSE \\
\midrule

\textbf{O2 Ground Truth:} & $16.0 \sum (\lambda_k^{-3} - 1) + 8.0 \sum (\lambda_k^{3} - 1)$ & \multicolumn{2}{c}{\bfseries ------} \\
0\% noise (LASSO-CV, 2) & $16.0 \sum (\lambda_k^{-3} - 1) + 8.0 \sum (\lambda_k^{3} - 1)$ & 1.0000 & 0.0000 \\
5\% noise (OMP-CV, 2) & $16.1 \sum (\lambda_k^{-3} - 1) + 7.9 \sum (\lambda_k^{3} - 1)$ & 0.9999 & 0.0018 \\
10\% noise (LASSO-AIC, 2) & $15.9 \sum (\lambda_k^{-3} - 1) + 8.0 \sum (\lambda_k^{3} - 1)$ & 0.9999 & 0.0017 \\
\midrule

\textbf{MR2 Ground Truth:} & $40.0 (I_1 - 3) + 20.0 (I_2 - 3)$ & \multicolumn{2}{c}{\bfseries ------} \\
0\% noise (LASSO-CV, 2) & $40.0 (I_1 - 3) + 20.0 (I_2 - 3)$ & 1.0000 & 0.0000 \\
5\% noise (LASSO-AIC, 2) & $39.8 (I_1 - 3) + 20.1 (I_2 - 3)$ & 1.0000 & 0.0009 \\
10\% noise (LASSO-CV, 2) & $40.3 (I_1 - 3) + 19.8 (I_2 - 3)$ & 0.9999 & 0.0019 \\
\midrule

\textbf{MR1O1 Ground Truth:} & $40.0 (I_2 - 3) + 8.0 \sum (\lambda_k^{-3} - 1)$ & \multicolumn{2}{c}{\bfseries ------} \\
0\% noise (LASSO-CV, 2) & $40.0 (I_2 - 3) + 8.0 \sum (\lambda_k^{-3} - 1)$ & 1.0000 & 0.0000 \\
5\% noise (LASSO-AIC, 2) & $40.4 (I_2 - 3) + 8.0 \sum (\lambda_k^{-3} - 1)$ & 1.0000 & 0.0010 \\
10\% noise (LASSO-BIC, 2) & $40.3 (I_2 - 3) + 8.0 \sum (\lambda_k^{-3} - 1)$ & 1.0000 & 0.0007 \\
\midrule

\textbf{MR2O2 Ground Truth:} & $40.0 (I_1 - 3) + 20.0 (I_2 - 3) + 16.0 \sum (\lambda_k^{-3} - 1) + 800.0 \sum (\lambda_k^{1} - 1)$ & \multicolumn{2}{c}{\bfseries ------} \\
0\% noise (LASSO-AIC, 4) & $40.0 (I_1 - 3) + 20.0 (I_2 - 3) + 16.0 \sum (\lambda_k^{-3} - 1) + 800.0 \sum (\lambda_k^{1} - 1)$ & 1.0000 & 0.0000 \\
5\% noise (LASSO-BIC, 4) & $42.8 (I_1 - 3) + 22.4 (I_2 - 3) + 16.2 \sum (\lambda_k^{-3} - 1) + 767.0 \sum (\lambda_k^{1} - 1)$ & 0.9994 & 0.0037 \\
10\% noise (LASSO-AIC, 4) & $49.8 (I_1 - 3) + 23.8 (I_2 - 3) + 15.5 \sum (\lambda_k^{-3} - 1) + 735.0 \sum (\lambda_k^{1} - 1)$ & 0.9990 & 0.0059 \\

\bottomrule
\end{tabular}%
}\\
\footnotesize{\textit{Note:} Ground truth \ac{SEF} coefficients are in Pa. Discovered model Ogden-type terms follow the form $\sum c_i(\sum_j \lambda_j^{\alpha_i} - 1)$. Discovered coefficients $c_i$ are from Ridge refit for the selected best-performing method (indicated in parentheses). $R^2_{\min}$ is $\min\{R^2(\text{UT } P_{11}), R^2(\text{PS } P_{11}), R^2(\text{EBT } P_{11})\}$. AvgNRMSE is the average of NRMSE values for the same three components.}
\end{table}

\begin{figure}[!htb]
\centering
\includegraphics[width=0.98\textwidth]{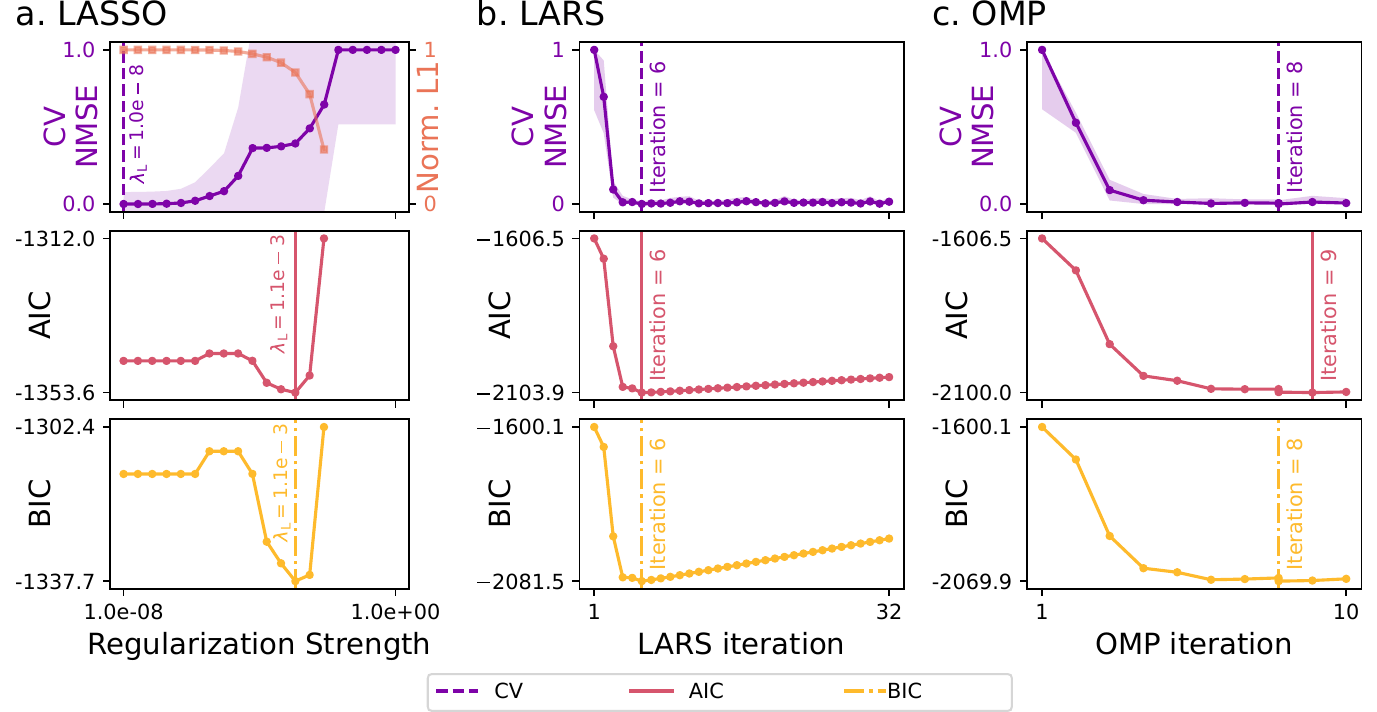}
    \caption{\textbf{Isotropic Synthetic Data:}.  
    Model selection for sparse regression for MR2O2 benchmark at 10\% relative noise. \textbf{a.} \ac{LASSO}, \textbf{b.} \ac{LARS}, and \textbf{c.} \ac{OMP}. Each row displays a different selection criterion metric against either the regularization penalty term (for \ac{LASSO}) or the number of steps (for \ac{LARS} and \ac{OMP}). Vertical dashed lines indicate the optimal complexity chosen by each criterion. The shaded area denotes the one-standard-error band for the \ac{CV} curve calculated for the different \ac{CV}-folds. NMSE stands for the normalized mean squared error.}
    \label{fig:relnoise10pct_synth_error_criteria}
\end{figure}

To gain a deeper insight into the discovery process, we analyze the behavior of different sparse regression algorithms and selection criteria for a representative case: the MR2O2 model with 10\% noise. Figure~\ref{fig:relnoise10pct_synth_error_criteria} visualizes the decision-making process for \ac{LASSO}, \ac{LARS}, and \ac{OMP}. For \ac{LASSO}, the \ac{CV}-NMSE is plotted against the regularization strength. The first column shows that, for \ac{LASSO}, \ac{CV} selects a model corresponding to the lowest regularization strength, while both the \ac{AIC} and \ac{BIC} criteria, plotted on their own axes, select a higher penalty; see Figure~\ref{fig:relnoise10pct_synth_error_criteria}a. 
For \ac{LARS} and \ac{OMP}, all selection criteria yield a similar result as seen in Figure~\ref{fig:relnoise10pct_synth_error_criteria}, columns b and c.
Despite differences between \ac{LASSO}, \ac{LARS}, and \ac{OMP}, the discovered models correctly resolve to the four ground truth terms in all cases after the final parameter refinement step; see Table~\ref{tab:synth_best_model_forms_kpa_nrmse_rel}.

\begin{figure}[!htb]
    \centering
    \includegraphics[width=0.8\textwidth]{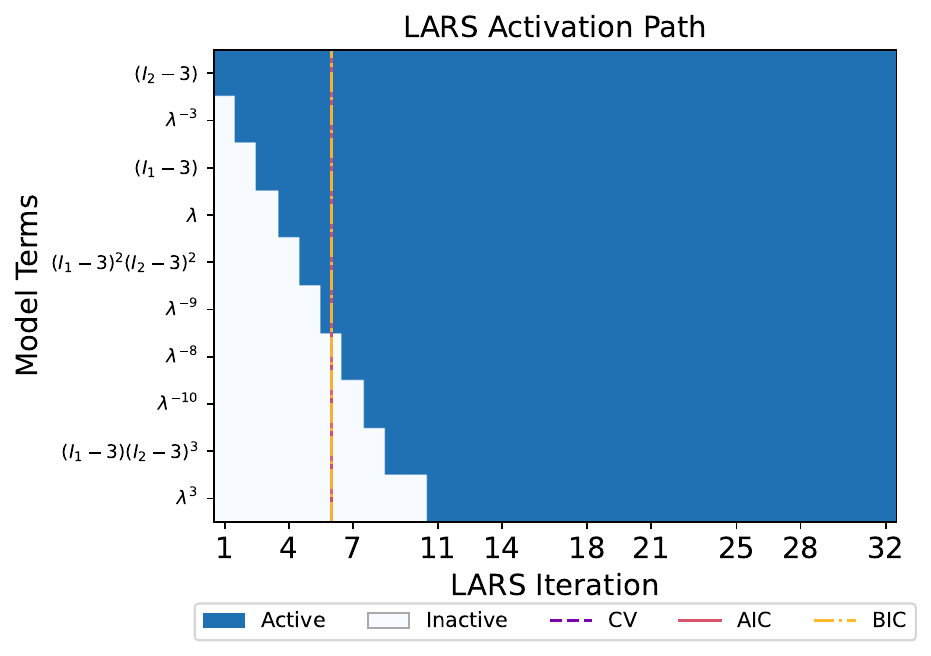}
    \caption{\textbf{Isotropic Synthetic Data:} \ac{LARS} activation path for MR2 benchmark at 10\% relative noise. The heatmap shows which model terms (y-axis) are active (blue) at each step of the algorithm (x-axis). The path illustrates the order in which terms are added to or removed from the model. Vertical lines mark the models selected by \ac{CV}, \ac{AIC}, and \ac{BIC}.}
    \label{fig:lars_activation_synthetic}
\end{figure}

\begin{figure}[!htb]
    \centering
    \includegraphics[width=0.8\textwidth]{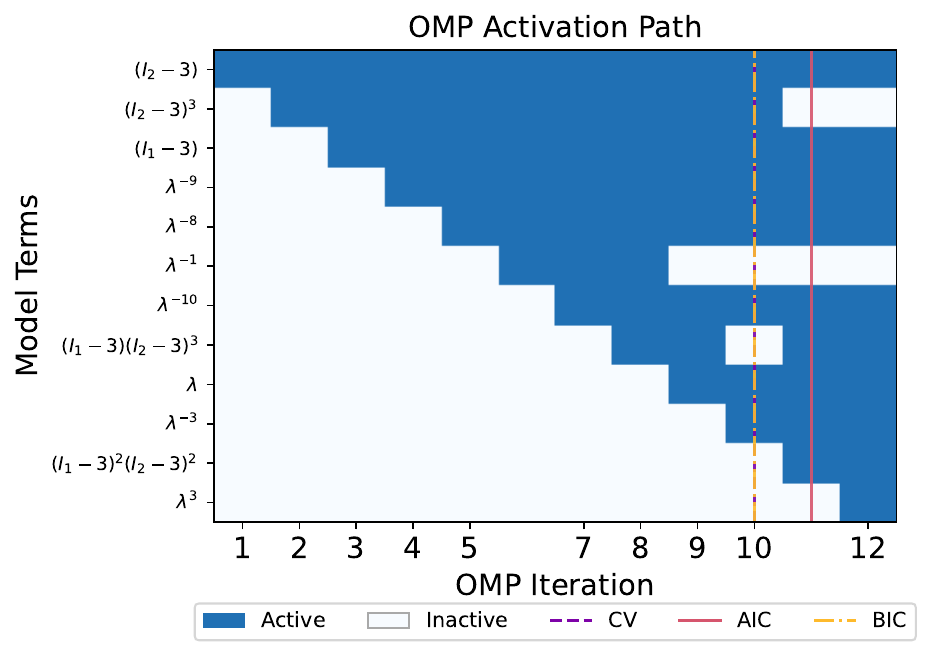}
    \caption{\textbf{Isotropic Synthetic Data:} \ac{OMP} activation path for MR2O2 benchmark at 10\% relative noise. The heatmap shows which model terms (y-axis) are active (blue) at each step of the algorithm (x-axis). The path illustrates the order in which terms are added to or removed from the model. Vertical lines mark the models selected by \ac{CV}, \ac{AIC}, and \ac{BIC}.}
    \label{fig:omp_activation_synthetic}
\end{figure}

The primary advantage of the forward selection methods, \ac{LARS} and \ac{OMP}, is their ability to provide a clear ranking of feature importance as the model is constructed step by step. This is illustrated in the corresponding activation paths for \ac{LARS} and \ac{OMP} in Figures~\ref{fig:lars_activation_synthetic} and \ref{fig:omp_activation_synthetic}, respectively. For \ac{LARS}, all three selection criteria (\ac{CV}, \ac{AIC}, and \ac{BIC}) initially agree on a six-term model. In contrast, for the \ac{OMP} algorithm, there is disagreement: \ac{CV} and \ac{BIC} select an eight-term model, while \ac{AIC} selects a nine-term model. The activation paths confirm that for \ac{LARS}, the correct ground truth terms, $I_1-3$, $I_2-3$, $\lambda^{-3}$, and $\lambda^{1}$, are among the very first to be activated, highlighting its capacity to rapidly identify dominant model features. In contrast, \ac{OMP} identifies the dominant Mooney-Rivlin terms within the first iterations but struggles to determine the correct Ogden terms, which highlights its greedy nature.
As with \ac{LASSO}, the final refitting stage is highly effective, and all \ac{LARS} and \ac{OMP} variants ultimately resolve to the correct four ground truth terms; see Table~\ref{tab:synth_best_model_forms_kpa_nrmse_rel}.

\subsection{Isotropic Experimental Data - Treloar Benchmark}\label{subsec:Treloar}

Having validated the framework's capability to recover known constitutive laws from synthetic data, we now apply it to the discovery of a constitutive model from the experimental Treloar dataset \cite{treloar_vulcanisedRubberData_1944}. 
The objective is to identify a model that is not only parsimonious but also accurately captures all three deformation modes present in the dataset simultaneously. This goal has been reported in the literature to be challenging \cite{ricker_systematicFittingHyperelasticity_2023}.
The model library is composed of 15 terms combining Mooney-Rivlin up to third-order and Ogden terms with $\alpha^{(l)} \in \lbrace -4, -3, -1, 1, 3, 4 \rbrace$. We omit the Ogden terms with exponents $\alpha^{(l)} \in \{-2, 2\}$ since they correspond to the Mooney-Rivlin terms
$\phi^{\textrm{MR}}_{(0,1)}$ and $\phi^{\textrm{MR}}_{(1,0)}$, respectively.

The results in Table~\ref{tab:performance_summary_treloar} show that the framework identifies several high-performing, physically consistent models. A notable finding is the strong consensus among different algorithms and selection criteria. All variants of the \ac{LASSO} and \ac{OMP} algorithms converge to the exact same compact four-term model. Likewise, all variants of the \ac{LARS} algorithm identify a second, distinct four-term model ($I_1$ and $\lambda_k^{-1}$ are replaced by $I_2$ and $\lambda_k^3$). Both discovered models achieve an excellent fit, with $R_{\min}^2=0.988$ in all three deformation modes (\ac{UT}, \ac{PS}, \ac{EBT}). This demonstrates the framework's ability to strike a balance between fidelity and parsimony when fitting to experimental data.

% \begin{table}[!hpbt]
% \centering
% \caption{\textbf{Isotropic Experimental Data:} Performance summary for models discovered from the Treloar dataset, evaluated on the \acf{UT}, \acf{PS}, and \acf{EBT} modes.}
% \label{tab:performance_summary_treloar}
% \begin{tabular}{@{}llcccccc@{}}
% \toprule
% {Algorithm} & {Selection} & {$n^{\mathcal{A}}_{\phi}$} & {R$_{UT}^2$} & {R$_{PS}^2$} & {R$_{EBT}^2$} & {AvgRMSE} & {Time [s]} \\
% & {criteria} & & & & & {[MPa]} & \\ \midrule
% {LASSO} & CV & 4 & 0.996 & 0.992 & 0.999 & 0.0555 & 0.4232 \\
% & AIC & 4 & 0.996 & 0.992 & 0.999 & 0.0555 & 0.0083 \\
% & BIC & 4 & 0.996 & 0.992 & 0.999 & 0.0555 & 0.0083 \\ \addlinespace
% {LARS} & CV & 4 & 0.995 & 0.988 & 0.997 & 0.0700 & 0.0026 \\
% & AIC & 4 & 0.995 & 0.988 & 0.997 & 0.0700 & 0.0026 \\
% & BIC & 4 & 0.995 & 0.988 & 0.997 & 0.0700 & 0.0026 \\ \addlinespace
% {OMP} & CV & 4 & 0.996 & 0.992 & 0.999 & 0.0555 & 0.0024 \\
% & AIC & 4 & 0.996 & 0.992 & 0.999 & 0.0555 & 0.0024 \\
% & BIC & 4 & 0.996 & 0.992 & 0.999 & 0.0555 & 0.0024 \\
% \bottomrule
% \end{tabular}
% \vspace{1ex}
% \begin{minipage}{0.9\textwidth}
% \footnotesize
% \textit{Notes:} {$n^{\mathcal{A}}_{\phi}$ indicates the number of active model library terms after refinement. R$^2$ values are reported for the primary stress component $P_{11}$. AvgRMSE is the averaged root mean squared error across all components of the dataset.}
% \end{minipage}
% \end{table}

%%% update 2025.11.19
\begin{table}[!hpbt]
\centering
\caption{\textbf{Isotropic Experimental Data:} Performance summary for models discovered from the Treloar dataset, evaluated on the \acf{UT}, \acf{PS}, and \acf{EBT} modes.}
\label{tab:performance_summary_treloar}
\begin{tabular}{@{}llcccccc@{}}
\toprule
{Algorithm} & {Selection} & {$n^{\mathcal{A}}_{\phi}$} & {R$_{UT}^2$} & {R$_{PS}^2$} & {R$_{EBT}^2$} & {AvgRMSE} & {Time [s]} \\
& {criteria} & & & & & {[MPa]} & \\ \midrule
{LASSO} & CV & 4 & 0.996 & 0.992 & 0.999 & 0.0632 & {0.3946} \\
& AIC & 4 & 0.996 & 0.992 & 0.999 & 0.0632 & {0.0072} \\
& BIC & 4 & 0.996 & 0.992 & 0.999 & 0.0632 & {0.0072} \\ \addlinespace
{LARS} & CV & 4 & 0.995 & 0.988 & 0.997 & 0.0669 & {0.0023} \\
& AIC & 4 & 0.995 & 0.988 & 0.997 & 0.0669 & {0.0023} \\
& BIC & 4 & 0.995 & 0.988 & 0.997 & 0.0669 & {0.0023} \\ \addlinespace
{OMP} & CV & 4 & 0.996 & 0.992 & 0.999 & 0.0632 & {0.0023} \\
& AIC & 4 & 0.996 & 0.992 & 0.999 & 0.0632 & {0.0023} \\
& BIC & 4 & 0.996 & 0.992 & 0.999 & 0.0632 & {0.0023} \\
\bottomrule
\end{tabular}
\vspace{1ex}
\begin{minipage}{0.9\textwidth}
\footnotesize
\textit{Notes:} {$n^{\mathcal{A}}_{\phi}$ indicates the number of active model library terms after refinement. R$^2$ values are reported for the primary stress component $P_{11}$. AvgRMSE is the averaged root mean squared error across all components of the dataset.}
\end{minipage}
\end{table}

The explicit forms of the discovered models (Table~\ref{tab:explicit_discovered_W_summary_treloar}) reveal a recurrent structure: a polynomial in the first or second invariant, $I_1$ or $I_2$, supplemented by a small number of Ogden-type terms. This mirrors the structure of established hyperelastic models \cite{steinmann_hyperelasticModelsTreloar_2012,holzapfel_nonlinearSolidMechanics_2000}, which captures the dominant entropic network response through an $I_1$ or $I_2$-polynomial and augments it with additional terms to model strain stiffening at large deformations. The data-driven selection of Ogden terms can thus be interpreted as an automated refinement of a physically meaningful baseline model.

\begin{table}[p]
\centering
\caption{\textbf{Isotropic Experimental Data:} Explicit form and performance metrics of the best-performing discovered \ac{SEF} ($\overline{W}$). The models were fitted to the Treloar dataset. Coefficients have units of MPa.}
\label{tab:explicit_discovered_W_summary_treloar}
\resizebox{0.97\textwidth}{!}{%
\small
\begin{tabularx}{\textwidth}{llXcc}
\toprule
{Algorithm} & {Selection} & {Explicit SEF Form ($\overline{W}$)} & {$R_{\min}^2$} & {AvgRMSE} \\
& criteria & &  & {[MPa]} \\
\midrule
\begin{tabular}[c]{@{}l@{}}{LASSO}/\\ {OMP}\end{tabular} & CV/AIC/BIC &
$\begin{aligned}[t]
\overline{W} & = \num{0.0752} (I_1-3) + \num{3.1e-5} (I_1-3)^3 \\
  & + \num{0.0819} \sum (\lambda_k^{-1}-1) + \num{0.4398} \sum (\lambda_k^{1}-1)
\end{aligned}$ & 0.992 & 0.0632 \\
\addlinespace
LARS & CV/AIC/BIC &
$\begin{aligned}[t]
\overline{W} & = \num{0.0024} (I_2-3) + \num{2.8e-5} (I_1-3)^3 \\
& + \num{0.7885} \sum (\lambda_k^{1}-1) + \num{0.0067} \sum (\lambda_k^{3}-1)
\end{aligned}$ & 0.988 & 0.0669 \\
\bottomrule
\end{tabularx}%
}
\end{table}

\begin{figure}[p]
    \centering
    \includegraphics[width=0.8\textwidth]{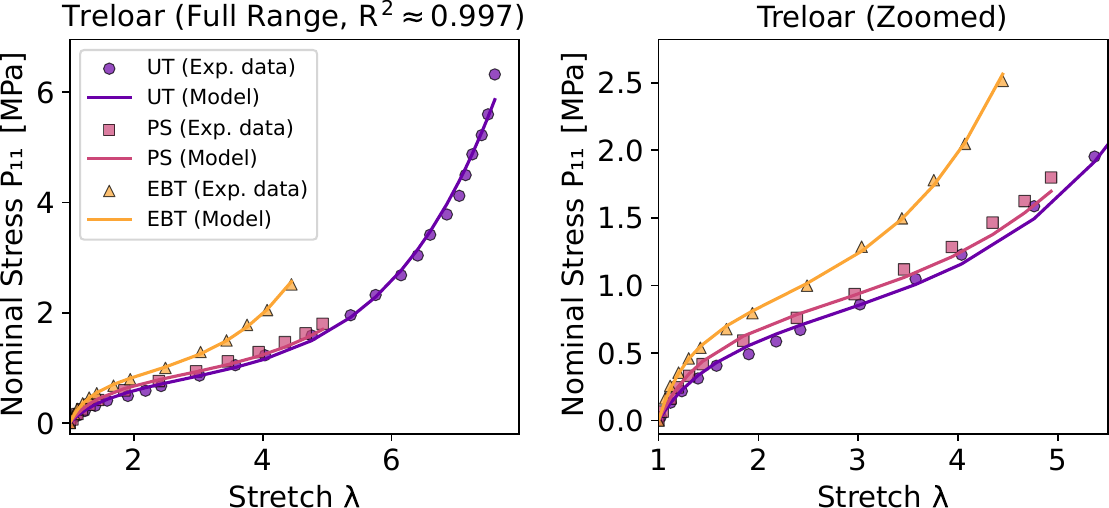}
    \caption{\textbf{Isotropic Experimental Data - \ac{LASSO}/\ac{OMP} Model:} Stress-stretch response of the four-term model discovered by \ac{LASSO} and \ac{OMP} methods compared against the Treloar experimental data. The model demonstrates excellent fidelity across the \acf{UT}, \acf{PS}, and \acf{EBT} deformation modes, confirming its high predictive accuracy. The reported $R^2$ value in this figure corresponds to the overall value for the three deformation modes.}
    \label{fig:lassobic_fit}
\end{figure}

For the experimental Treloar dataset, the choice of model selection criterion has a minimal effect on the discovered model. In this analysis, all selection criteria, \ac{CV}, \ac{AIC}, and \ac{BIC}, demonstrated remarkable consistency within each algorithm family. The performance of the four-term model discovered by the \ac{LASSO} and \ac{OMP} methods is illustrated in Figure~\ref{fig:lassobic_fit}, showing excellent agreement with the Treloar dataset across all deformation modes.

\begin{figure}[p]
    \centering
    \includegraphics[width=0.98\textwidth]{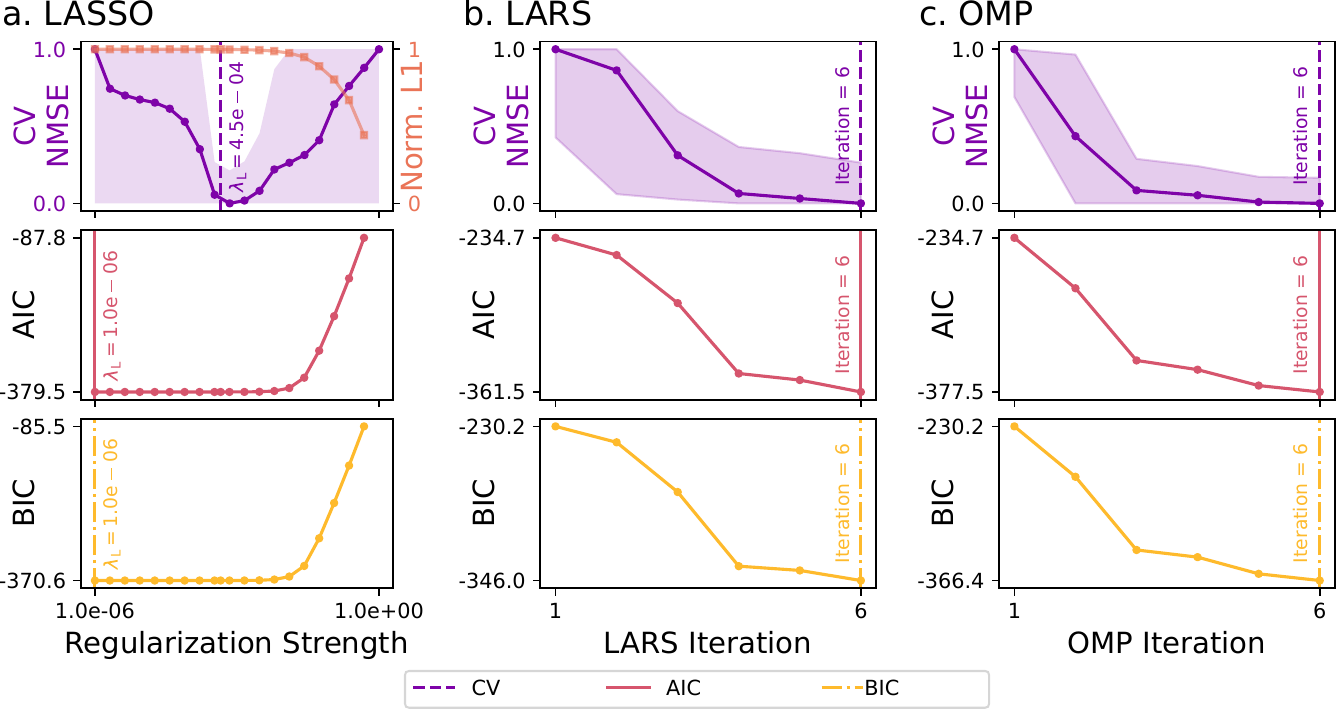}
    \caption{\textbf{Isotropic Experimental Data:} Model selection for sparse regression on the Treloar dataset. \textbf{a.} \ac{LASSO}, \textbf{b.} \ac{LARS}, and \textbf{c.} \ac{OMP}. Each row displays a different selection criterion metric against either the regularization penalty term (for \ac{LASSO}) or the number of terms (for \ac{LARS} and \ac{OMP}). Vertical dashed lines indicate the optimal complexity chosen by each criterion. The shaded area denotes the one-standard-error band for the \ac{CV} curve calculated for the different \ac{CV}-folds. NMSE stands for the normalized mean squared error.}
    \label{fig:treloar_selection_criteria}
\end{figure}

Figure~\ref{fig:treloar_selection_criteria} displays the \ac{CV} error and \ac{AIC}/\ac{BIC} values for \ac{LASSO}, \ac{LARS}, and \ac{OMP}. It is observed from the figure that the regularization strength at the minimum \ac{CV} error and \ac{AIC}/\ac{BIC} is selected. \Ac{LARS} and \ac{OMP} for all model selection criteria stop after six iterations.

\begin{figure}[p]
    \centering
    \includegraphics[width=0.8\textwidth]{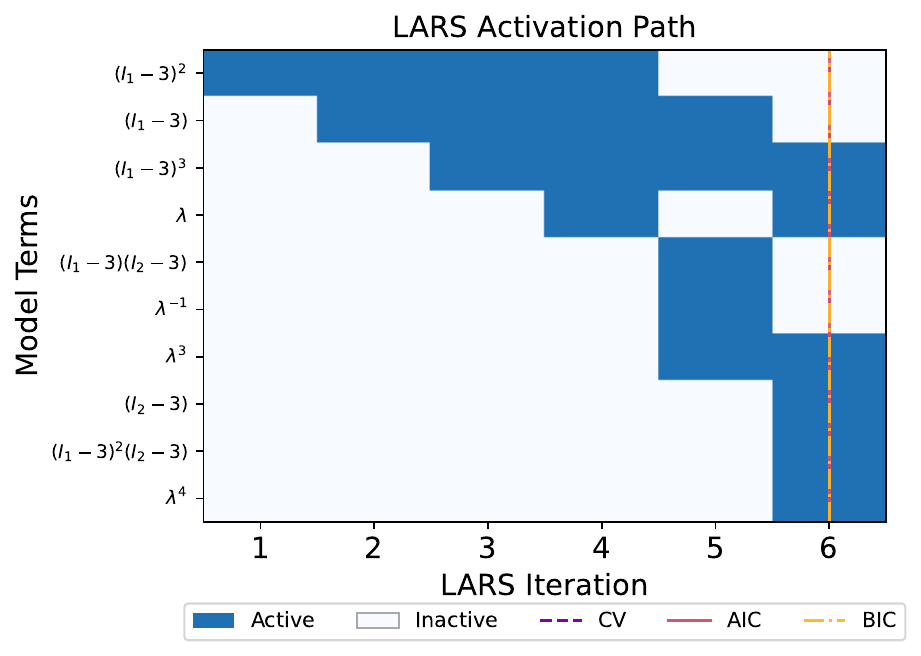}
    \caption{\textbf{Isotropic Experimental Data - Activation Paths:} Visualization of the forward selection process for \ac{LARS}. The heatmap shows which model terms (y-axis) are active (blue) at each step of the algorithm (x-axis). The path illustrates the order in which terms are added to or removed from the model. Vertical lines mark the models selected by \ac{CV}, \ac{AIC}, and \ac{BIC}.}
    \label{fig:treloar_lars_activation_path}
\end{figure}

\begin{figure}[p]
    \centering
    \includegraphics[width=0.8\textwidth]{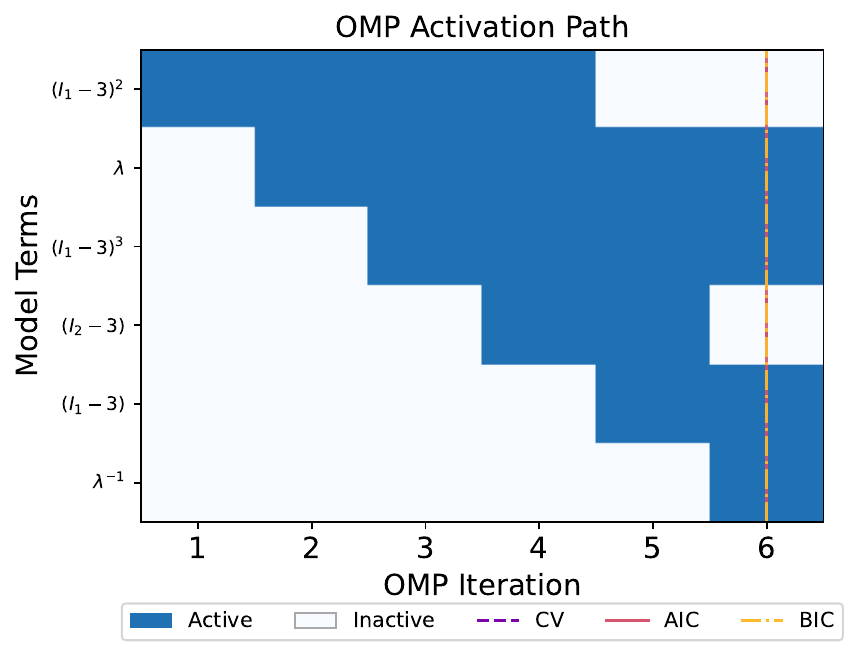}
    \caption{\textbf{Isotropic Experimental Data - Activation Paths:} Visualization of the forward selection process for \ac{OMP}. The heatmap shows which model terms (y-axis) are active (blue) at each step of the algorithm (x-axis). The path illustrates the order in which terms are added to or removed from the model. Vertical lines mark the models selected by \ac{CV}, \ac{AIC}, and \ac{BIC}.}
    \label{fig:treloar_omp_activation_path}
\end{figure}

The activation paths in Figures~\ref{fig:treloar_lars_activation_path} and \ref{fig:treloar_omp_activation_path} offer additional insight into the distinct discovery strategies of the algorithms. Both the \ac{LARS} path in Figure~\ref{fig:treloar_lars_activation_path} and the \ac{OMP} path in Figure~\ref{fig:treloar_omp_activation_path} confirm a structured model construction. This pathwise approach identifies and adds terms based on their evolving correlation with the model residuals, often revealing underlying physical relationships in the order of selection. Interestingly, in both cases, the Mooney-Rivlin term $(I_1 - 3)^2$ is initially marked as the most descriptive term but is subsequently replaced by terms that performed equally well. It is also observed that \ac{LARS} considers some of the mixed Mooney-Rivlin terms, while \ac{OMP} includes none of them. Actually, while \ac{LARS} explores the suitability of ten different terms, \ac{OMP} efficiently discovers the four-term model after exploring the combination of only six different terms. 

\subsection{Anisotropic Experimental Data - Human Cardiac Tissue}
\label{subsec:anisotropic_analysis_final}

Having demonstrated the framework's capabilities on isotropic materials, we now address the discovery of an anisotropic constitutive model from experimental data of human cardiac tissue \cite{martonová_modelDiscoveryCardiacTissue_2025}. This problem is substantially more complex due to the need to capture directionally dependent mechanical responses. Therefore, a model library with non-linear parameters is needed. Gaussian noise, with standard deviations proportional to the experimental stress values (5\% and 10\% of the true stress), was added to the stress components to evaluate the robustness of the nine algorithms against noise.

Martonová et al. (2024) \cite{martonová_modelDiscoveryCardiacTissue_2025} discovered a 4-term model under 3\% noise from experimental data of human cardiac tissue by solving the $\ell_1$-penalized version of \textbf{Problem 2} \eqref{eq:l1_constrained_problem_abstract}, with $\lambda_L = 0.01$.
The explicit form of the 4-term discovered model is reported in Table~\ref{tab:aniso_model_comparison_final_v2}. We use this 4-term model as a baseline for comparison purposes with respect to the model discovery algorithms proposed in this work.
This 4-term model yields an overall $R^2$ of 0.851 and an \ac{RMSE} of 0.523. In \cite{martonová_modelDiscoveryCardiacTissue_2025}, it was reported that 3 to 8 hours were required to solve the non-linear optimization problem.

To enable the use of \ac{LASSO}, \ac{LARS}, and \ac{OMP} as introduced in Section~\ref{sec:Methodology}, we temporarily fixed all non-linear inner coefficients of the exponential terms to unity, i.e., $\vecw = \bm{1}$, when assembling the linear design matrix in Eq.~\eqref{eq:linearized_design_matrix}. 
This choice for $\vecw$ was made such that the stress contribution of the exponential terms remains on the same order of magnitude as the linear and quadratic terms, but fixing $\vecw$ at a different value is also possible. We observed that $\vecw \in [0.1, 10]$ yielded nearly identical results when solving the sparse regression problem. Note that if $\vecw$ goes to zero, the exponential goes to one, and the overall stress contribution becomes a constant. In contrast, for larger values of $\vecw$, the exponential terms grow too rapidly, causing some numerical instabilities.  
The linearization of the design matrix facilitates the rapid identification of the most promising model terms that describe the anisotropic material response. 

\begin{figure}[p]
\centering
\includegraphics[width=0.98\textwidth]{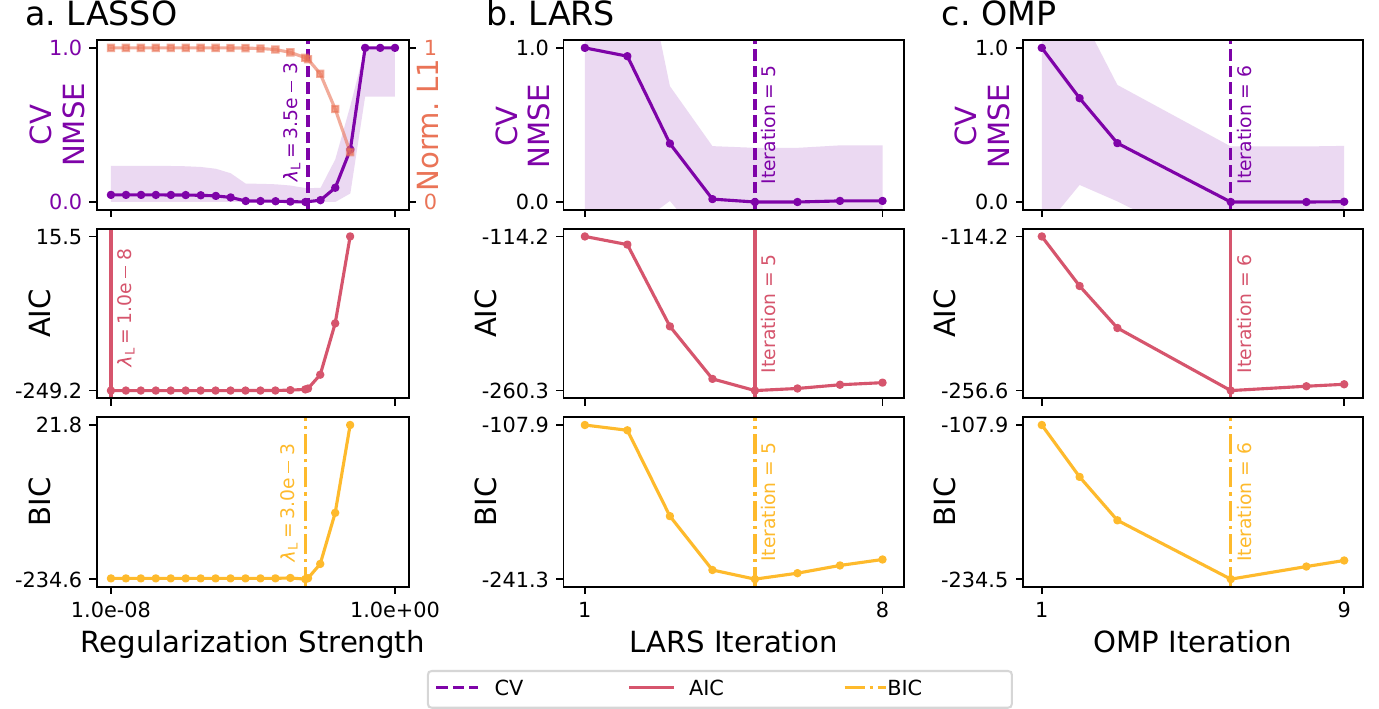}
\caption{\textbf{Anisotropic Experimental Data:} Model selection via sparse regression on data with 10\% noise. \textbf{a.} \ac{LASSO}, \textbf{b.} \ac{LARS}, and \textbf{c.} \ac{OMP}. Each row displays a different selection criterion metric against either the regularization penalty term (for \ac{LASSO}) or the number of active terms (for \ac{LARS} and \ac{OMP}). Vertical dashed lines indicate the optimal complexity chosen by each criterion. The shaded area denotes the one-standard-error band for the \ac{CV} curve calculated for the different \ac{CV}-folds. NMSE stands for the normalized mean squared error.}
\label{fig:relnoise10pct_stage1_criteria}
\end{figure}

The model selection process for the 10\% noise scenario (Figure~\ref{fig:relnoise10pct_stage1_criteria}) demonstrates the framework's effectiveness. For the 0\% and 5\% noise cases, 4-term models were discovered, while at 10\% noise, several methods converged to 5-term models. All of the discovered models have higher $R^2$ values and lower \ac{RMSE} than the baseline; refer to Table~\ref{tab:aniso_performance_summary_final_v2_with_times}.

\begin{figure}[!htb]
    \centering
    \includegraphics[width=0.8\textwidth]{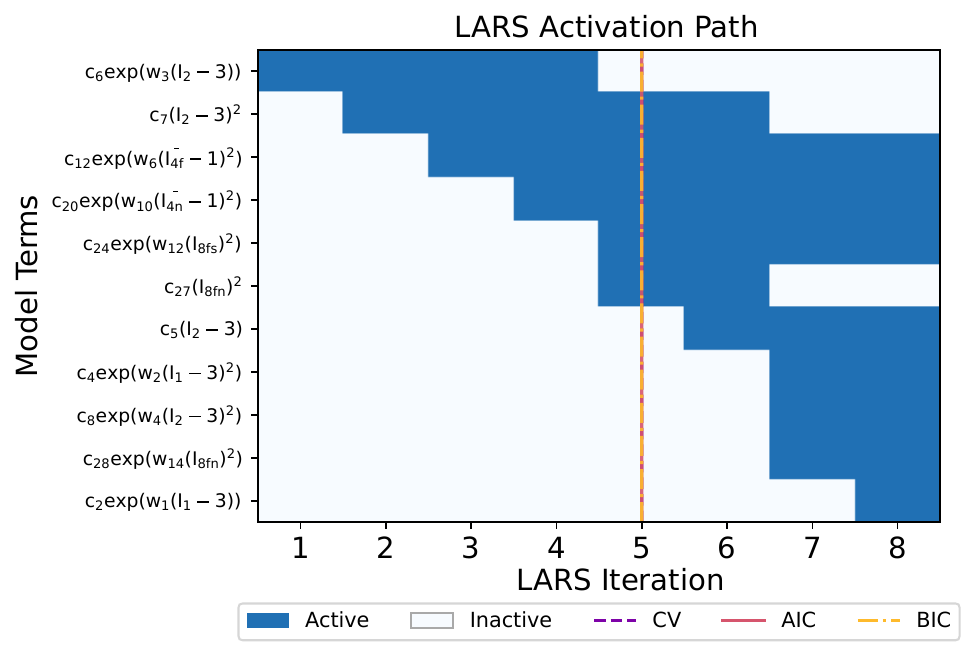}
    \caption{\textbf{Anisotropic Experimental Data:} 
    Visualization of the forward selection process for \ac{LARS}. The heatmap shows which model terms (y-axis) are active (blue) at each step of the algorithm (x-axis). The path illustrates the order in which terms are added to or removed from the model. Vertical lines mark the models selected by \ac{CV}, \ac{AIC}, and \ac{BIC}.}
    \label{fig:lars_activation}
\end{figure}

\begin{figure}[!htb]
    \centering
    \includegraphics[width=0.8\textwidth]{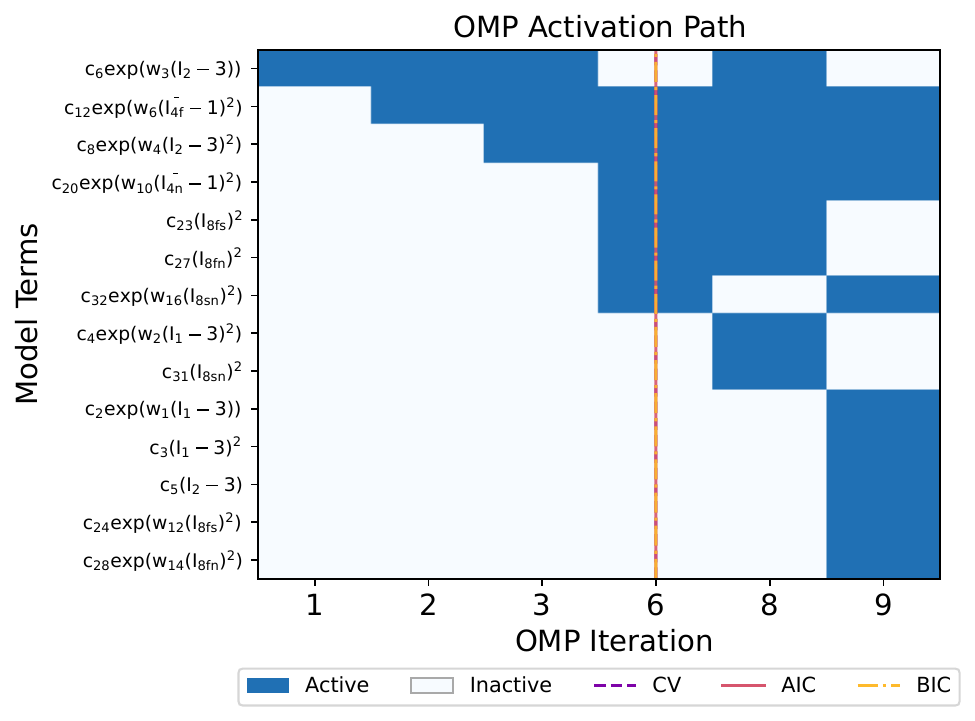}
    \caption{\textbf{Anisotropic Experimental Data:} Visualization of the forward selection process for \ac{OMP}. The heatmap shows which model terms (y-axis) are active (blue) at each step of the algorithm (x-axis). The path illustrates the order in which terms are added to or removed from the model. Vertical lines mark the models selected by \ac{CV}, \ac{AIC}, and \ac{BIC}.}
    \label{fig:omp_activation}
\end{figure}

The activation paths for \ac{LARS} and \ac{OMP} (Figures~\ref{fig:lars_activation} and \ref{fig:omp_activation}) reveal that both algorithms prioritize the isotropic ground matrix response, initially selecting terms related to the invariant $I_2$ to capture the baseline tissue behavior. The refinement stage consistently finalizes the isotropic contribution with either a quadratic polynomial (term~7, $c_7[I_2-3]^2$) or its quadratic exponential variant (term~8, $c_8\exp{w_4(I_2-3)^2}$). This indicates that while the linearized selection stage identifies the correct {invariant} early, it cannot distinguish the optimal {functional form} until the non-linear coefficients are freed.

\begin{figure}[p]
\centering
\includegraphics[width=0.98\textwidth]{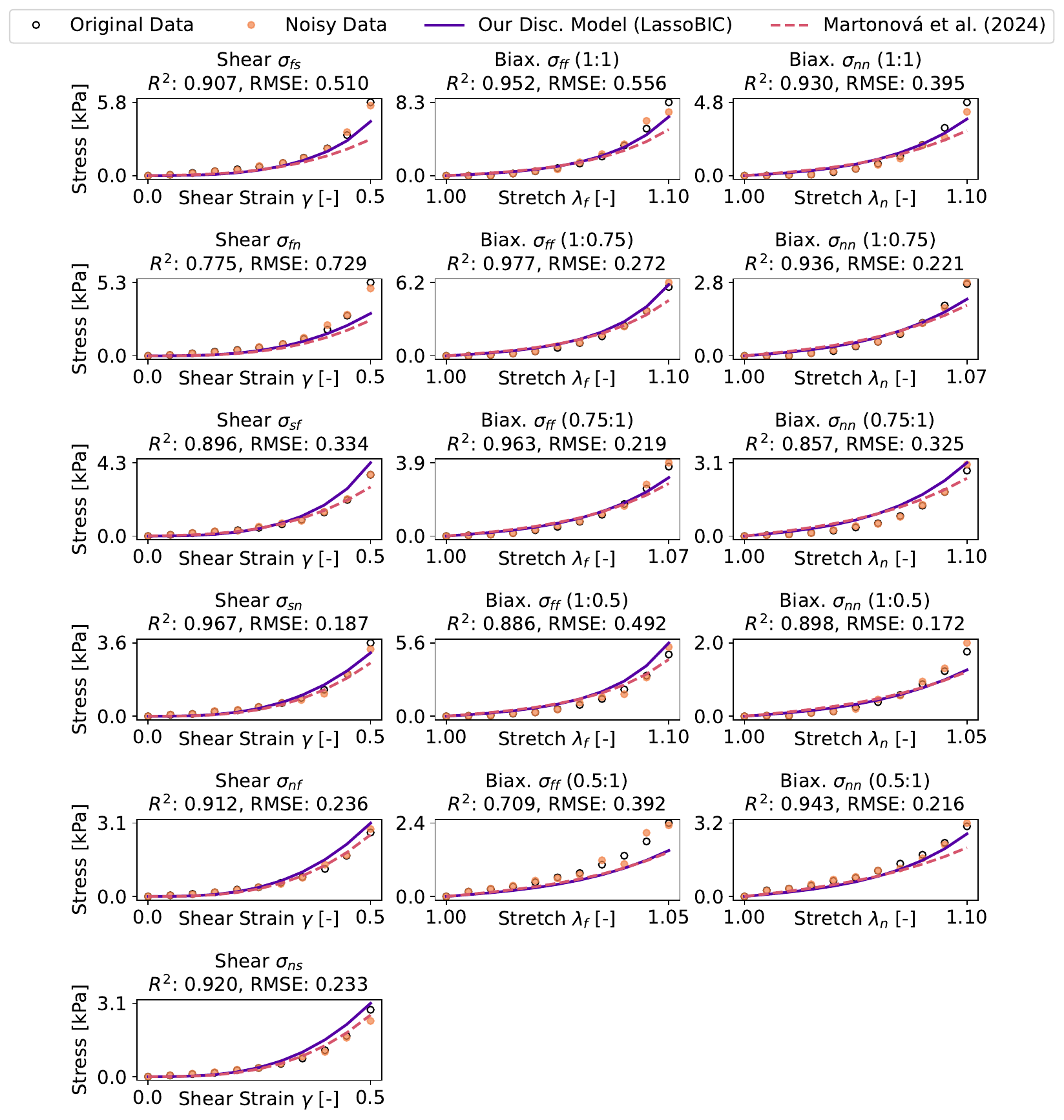}
\caption{\textbf{Anisotropic Experimental Data - \ac{LASSO}-\ac{BIC} at 10\% Noise:} Comparison of the discovered model's predictions against experimental data. Results for six triaxial shear tests and five biaxial extension protocols are shown. Experimental data points are shown for the original data (from Martonová et al. (2024) \cite{martonová_modelDiscoveryCardiacTissue_2025}, used as reference) and the 10\% noisy data used for model discovery. Predictions are shown for: \textbf{i.} the discovered 4-term anisotropic model obtained using \ac{LASSO}-\ac{BIC} at 10\% noise (solid lines), and \textbf{ii.} the baseline model (dashed lines). The $R^2$ and \ac{RMSE} values displayed in each subplot correspond to the performance of the \ac{LASSO}-\ac{BIC} model against the original experimental data for that specific deformation mode.}
\label{fig:lassobic_10noise_comparison}
\end{figure}

Finally, the discovery algorithms introduced here not only yielded more accurate results but also were more computationally efficient than the approach proposed in \cite{martonová_modelDiscoveryCardiacTissue_2025}, as detailed in Table~\ref{tab:aniso_performance_summary_final_v2_with_times}. For instance, the best-performing model at 5\% noise (\ac{LASSO}-\ac{CV}) was identified and refined in approximately 24 seconds. This represents a significant improvement in computational time with respect to Martonová et al (2024) \cite{martonová_modelDiscoveryCardiacTissue_2025}, making the framework a practical tool for the rapid discovery of constitutive models. 
The framework's effectiveness is best demonstrated by its performance under the most challenging conditions. 
Figure~\ref{fig:lassobic_10noise_comparison} shows the comparison of the discovered model by {LASSO-BIC} at 10\% noise with respect to the experimental data and Martonová et al. (2024) \cite{martonová_modelDiscoveryCardiacTissue_2025}.
The explicit \ac{SEF} form of the discovered model is reported in Table~\ref{tab:aniso_model_comparison_final_v2}.
Note that the discovered model by \ac{LASSO}-\ac{BIC} has the same functional form as the baseline model, but the value of the material parameters varies.
The model discovered by \ac{LASSO}-\ac{BIC} at 10\% noise improves the $R^2$ from the baseline's 0.851 to 0.923 and reduces the \ac{RMSE} from 0.523 to 0.376~kPa. 
This outcome illustrates how, for the discovery of anisotropic models, the discovery algorithms obtained by pairing sparse regression with selection criteria can consistently select compact yet highly accurate models. 

% --- TABLE: Anisotropic Performance Summary (Final, All-Inclusive) ---
\begin{table}[!htb]
\centering
\caption{\textbf{Anisotropic Experimental Data:} Performance summary for all discovered anisotropic models. $n^{\mathcal{A}}_{\phi}$ is the number of active terms after refinement. Time for sparse regression and refinement is given in seconds.}
\label{tab:aniso_performance_summary_final_v2_with_times}
\resizebox{\textwidth}{!}{%
\begin{tabular}{@{}lllccccc@{}}
\cmidrule(lr){4-8}
& & & $n^{\mathcal{A}}_{\phi}$ & Overall $R^2$ & \begin{tabular}[c]{@{}c@{}}Overall \\ RMSE (kPa)\end{tabular} & \multicolumn{2}{c}{Computation time} \\
\midrule
% \multicolumn{7}{l}{\textbf{Baseline}} \\
\multicolumn{3}{l}{\textbf{Martonová et al. (2024)}} & \textbf{4} & \textbf{0.851} & \textbf{0.523} & \multicolumn{2}{c}{$\sim$ \textbf{3 - 8} hours} \\
\midrule
% \multicolumn{7}{l}{\textbf{Selection and Refinement}} \\
\textbf{Noise level} & \textbf{Algorithm} & \begin{tabular}[c]{@{}l@{}}\textbf{Selection}\\ \textbf{criteria}\end{tabular} & & & & \textbf{Sparse Reg. {[s]}} & \textbf{Refinement {[s]}} \\
\cmidrule(lr){7-8}
\textbf{0\% noise} & \textbf{LASSO} & \textbf{CV} & \textbf{4} & \textbf{0.925} & \textbf{0.370} & \textbf{0.65} & \textbf{20.90} \\
& & AIC & 4 & 0.923 & 0.376 & 0.65 & 23.89 \\
& & BIC & 4 & 0.923 & 0.376 & 0.65 & 22.63 \\
& LARS & CV & 4 & 0.922 & 0.377 & 0.01 & 24.24 \\
& & AIC & 4 & 0.922 & 0.377 & 0.01 & 24.97 \\
& & BIC & 4 & 0.922 & 0.377 & 0.01 & 25.17 \\
& OMP & CV & 4 & 0.921 & 0.380 & 0.77 & 19.42 \\
& & AIC & 4 & 0.921 & 0.380 & 0.77 & 19.27 \\
& & BIC & 4 & 0.921 & 0.380 & 0.77 & 18.54 \\
\midrule
\textbf{5\% noise} & \textbf{LASSO} & \textbf{CV} & \textbf{4} & \textbf{0.925} & \textbf{0.371} & \textbf{0.56} & \textbf{24.05} \\
& & AIC & 4 & 0.923 & 0.375 & 0.56 & 20.86 \\
& & BIC & 4 & 0.923 & 0.375 & 0.56 & 21.90 \\
& LARS & CV & 4 & 0.923 & 0.375 & 0.01 & 22.87 \\
& & AIC & 4 & 0.923 & 0.375 & 0.01 & 21.64 \\
& & BIC & 4 & 0.923 & 0.375 & 0.01 & 20.01 \\
& OMP & CV & 4 & 0.921 & 0.380 & 0.36 & 17.27 \\
& & AIC & 4 & 0.921 & 0.380 & 0.36 & 17.37 \\
& & BIC & 4 & 0.921 & 0.380 & 0.36 & 18.04 \\
\midrule
\textbf{10\% noise} & LASSO & CV & 4 & 0.922 & 0.378 & 0.61 & 22.93 \\
& & AIC & 5 & 0.921 & 0.379 & 0.61 & 34.79 \\
& & \textbf{BIC} & \textbf{4} & \textbf{0.923} & \textbf{0.376} & \textbf{0.61} & \textbf{23.25} \\
& LARS & CV & 5 & 0.920 & 0.383 & 0.01 & 33.20 \\
& & AIC & 5 & 0.920 & 0.383 & 0.01 & 32.94 \\
& & BIC & 5 & 0.920 & 0.383 & 0.01 & 32.92 \\
& OMP & CV & 5 & 0.919 & 0.385 & 0.25 & 26.16 \\
& & AIC & 5 & 0.919 & 0.385 & 0.25 & 27.12 \\
& & BIC & 5 & 0.919 & 0.385 & 0.25 & 27.15 \\
\bottomrule
\end{tabular}%
}
\begin{minipage}{\textwidth}
\footnotesize
\textit{Notes:} The baseline model is the 4-term model discovered by Martonová et al. (2024) \cite{martonová_modelDiscoveryCardiacTissue_2025}; performance values are from our evaluation. The $R^2$ and \ac{RMSE} values correspond to the performance of the models against the original experimental data. The best-performing models at each noise level based on the overall \ac{RMSE} and $R^2$ are highlighted in bold.
\end{minipage}
\end{table}

\begin{table}[p]
\centering
\caption{\textbf{Anisotropic Experimental Data:} Explicit \ac{SEF} form ($\overline{W}$) for Martonová et al. (2024) \cite{martonová_modelDiscoveryCardiacTissue_2025} and the discovered model (\ac{LASSO}-\ac{BIC}) at 10\% relative noise. Coefficients have units of kPa.}
\label{tab:aniso_model_comparison_final_v2}
\resizebox{0.97\textwidth}{!}{%
\small
\begin{tabularx}{\textwidth}{llXcc}
\toprule
{Algorithm/} & {Selection} & {Explicit SEF form ($\overline{W}$)} & {$R^{2}$} & {RMSE} \\
 Noise level & criteria  &  &  & {[kPa]} \\
\midrule
\begin{tabular}[c]{@{}l@{}}\textbf{Baseline}\\ 3\% noise\end{tabular}
& --- &
$\begin{aligned}[t]
\overline{W} & = \num{5.1620}\,[I_2-3]^2 \\
    & + \num{0.0810}\,[\exp{\num{21.1510}\,(\max\{I_{\textrm{4f}},1}-1)^2\}-1] \\
    & + \num{0.3150}\,[\exp{\num{4.3710}\,(\max\{I_{\textrm{4n}},1}-1)^2\}-1] \\
    & + \num{0.4860}\,[\exp{\num{0.5080}\,I_{\textrm{8fs}}^2}-1]
\end{aligned}$ & 0.851 & 0.523 \\
\addlinespace
\begin{tabular}[c]{@{}l@{}}\textbf{LASSO}\\ 10\% noise\end{tabular} & BIC &
$\begin{aligned}[t]
\overline{W} & = \num{6.1424}\,[I_2 - 3]^2 \\
& + \num{0.0459}\,\bigl[\exp{\num{31.2897}\,(\max\{I_{\textrm{4f}},1\}-1)^2}-1\bigr] \\
& + \num{0.0661}\,\bigl[\exp{\num{16.3484}\,(\max\{I_{\textrm{4n}},1\}-1)^2}-1\bigr] \\
& + \num{0.0035}\,\bigl[\exp{\num{13.1169}\,I_{\textrm{8fs}}^{2}}-1\bigr]
\end{aligned}$ & 0.923 & 0.376 \\
\bottomrule
\end{tabularx}%
}
\end{table}

\afterpage{\clearpage}
\newpage

\section{Conclusions}\label{sec:conclusions} 

This work has presented a fully automated framework for constitutive model discovery, combining three classes of sparse regression algorithms (\ac{LASSO}, \ac{LARS}, and \ac{OMP}) with three model selection criteria (\ac{CV}, \ac{AIC}, and \ac{BIC}). 
The results show that all nine algorithm–criterion combinations perform consistently well. In synthetic benchmarks, the ground-truth models were reliably discovered even under 5\% and 10\% noise scenarios. For Treloar’s data, the framework consistently discovered four-term models that accurately capture the different deformation modes. In the case of anisotropic cardiac tissue, the same robustness was observed, with all pipelines producing parsimonious models that capture the essential features of the material response and performed comparatively better than a state-of-the-art baseline model.

These findings broaden the range of viable discovery algorithms. Beyond the well-established \ac{LASSO}, we demonstrate that both \ac{LARS} and \ac{OMP} are equally effective alternatives when coupled with model selection criteria, thereby enlarging the family of sparse regression methods available for constitutive model discovery. 
The proposed framework eliminates manual intervention from the model selection process while reducing the computational time to a fraction of that required by $\ell_1$-based discovery.
The overall framework thus provides a reliable basis for the automated discovery of models for both isotropic and anisotropic hyperelastic materials.

\newpage
\section*{CRediT authorship contribution statement}

\textbf{Jorge-Humberto Urrea-Quintero:} Conceptualization, Methodology, Software, Validation, Formal analysis, Investigation, Data Curation, Writing - Original Draft, Writing - Review \& Editing, Visualization
\textbf{David Anton:} Conceptualization, Methodology, Validation, Formal analysis, Investigation, Writing - Original Draft, Writing - Review \& Editing
\textbf{Laura De Lorenzis:} Conceptualization, Methodology, Writing - Review \& Editing, Supervision 
\textbf{Henning Wessels:} Conceptualization, Methodology, Resources, Writing - Original Draft, Writing - Review \& Editing, Supervision, Funding acquisition

\section*{Declaration of competing interest}
The authors declare that they have no known competing financial interests or personal relationships that could have appeared to influence the work reported in this paper.

\section*{Acknowledgments}
David Anton and Henning Wessels acknowledge support in the project DFG 501798687: \textit{"Monitoring data driven life cycle management with AR based on adaptive, AI-supported corrosion prediction for reinforced concrete structures under combined impacts"} which is a subproject of SPP 2388: \textit{"Hundred plus - Extending the Lifetime of Complex Engineering Structures through Intelligent Digitalization"} funded by the DFG. Henning Wessels also acknowledges BMUV 02E12244C: \textit{Verbundprojekt: KI-unterstützte Stoffmodellierung am Beispiel von Bentonit (KI-Stoff), Teilprojekt C}. Laura De Lorenzis acknowledges funding by SNF through grant 200021 204316 “Unsupervised data-driven discovery of material laws”.

\section*{Data availability}
Our research code is available on Zenodo \cite{urrea_codeAutomatedModelDiscovery_2025} and GitHub: \url{https://github.com/jhurreaq/MDisc_pairing.git}.

\newpage
\appendix
\section{Orthotropic Model Library}\label{sec:ortho_model}

For the special case of orthotropy, the preferred directions are encoded in the structural tensors $\tenM_{\mathrm{f}}$, $\tenM_{\mathrm{s}}$, and $\tenM_{\mathrm{n}}$. The isochoric \ac{SEF} $\overline{W}$ is expressed in terms of the isotropic invariants ($I_1, I_2$) and a set of anisotropic invariants derived from the right Cauchy-Green tensor $\tenC$. Assuming incompressibility ($J=1$), the principal anisotropic invariants are:
\begin{itemize}
    \item \textbf{Stretch-related invariants} ($I_4$-type):
    \begin{equation}\label{eq:inv_stretch_aniso}
    \begin{aligned}
        I_{\mathrm{4f}}(\tenC, \tenM_{\mathrm{f}}) &= \vecf_0 \cdot \tenC \vecf_0, \\
        I_{\mathrm{4s}}(\tenC, \tenM_{\mathrm{s}}) &= \vecs_0 \cdot \tenC \vecs_0, \\
        I_{\mathrm{4n}}(\tenC, \tenM_{\mathrm{n}}) &= \vecn_0 \cdot \tenC \vecn_0.
    \end{aligned}
    \end{equation}
    \item \textbf{Coupling invariants} ($I_8$-type):
    \begin{equation}\label{eq:coupling_inv_aniso}
    \begin{aligned}
        I_{\mathrm{8fs}}(\tenC, \tenM_{\mathrm{f}}, \tenM_{\mathrm{s}}) &= \vecf_0 \cdot \tenC \vecs_0, \\
        I_{\mathrm{8fn}}(\tenC, \tenM_{\mathrm{f}}, \tenM_{\mathrm{n}}) &= \vecf_0 \cdot \tenC \vecn_0, \\
        I_{\mathrm{8sn}}(\tenC, \tenM_{\mathrm{s}}, \tenM_{\mathrm{n}}) &= \vecs_0 \cdot \tenC \vecn_0.
    \end{aligned}
    \end{equation}
\end{itemize}
Based on these invariants, the 32-term orthotropic library from \cite{martonová_modelDiscoveryCardiacTissue_2025} is constructed from the following basis functions $\phi_{j}$. The non-linear parameters $w_{j}$ only exist for the exponential terms (even-numbered basis functions):
\begin{equation}
    \begin{aligned}
        (I_1 - 3) &\begin{cases}
            \phi_1 = (I_1-3)                & \phi_2 = \exp{w_{2}(I_1-3)}-1  \\
            \phi_3 = (I_1-3)^2               & \phi_4 = \exp{w_{4}(I_1-3)^2}-1
        \end{cases},\\
        (I_2 - 3)&\begin{cases}
            \phi_5 = (I_2-3)                & \phi_6 = \exp{w_{6}(I_2-3)}-1  \\
            \phi_7 = (I_2-3)^2               & \phi_8 = \exp{w_{8}(I_2-3)^2}-1
        \end{cases},\\     
        (\max\{I_{\textrm{4f}},1\}-1)&\begin{cases}
            \phi_9 = (\max\{I_{\textrm{4f}},1\}-1)      & \phi_{10} = \exp{w_{10}(\max\{I_{\textrm{4f}},1\}-1)}-1 \\ 
            \phi_{11} = (\max\{I_{\textrm{4f}},1\}-1)^2     & \phi_{12} = \exp{w_{12}(\max\{I_{\textrm{4f}},1\}-1)^2}-1
        \end{cases},\\     
        (\max\{I_{\textrm{4s}},1\}-1)&\begin{cases}
            \phi_{13} = (\max\{I_{\textrm{4s}},1\}-1)       & \phi_{14} = \exp{w_{14}(\max\{I_{\textrm{4s}},1\}-1)}-1 \\
            \phi_{15} = (\max\{I_{\textrm{4s}},1\}-1)^2      & \phi_{16} = \exp{w_{16}(\max\{I_{\textrm{4s}},1\}-1)^2}-1
        \end{cases},\\
        (\max\{I_{\textrm{4n}},1\}-1)&\begin{cases}
            \phi_{17} = (\max\{I_{\textrm{4n}},1\}-1)     & \phi_{18} = \exp{w_{18}(\max\{I_{\textrm{4n}},1\}-1)}-1 \\
            \phi_{19} = (\max\{I_{\textrm{4n}},1\}-1)^2      & \phi_{20} = \exp{w_{20}(\max\{I_{\textrm{4n}},1\}-1)^2}-1
        \end{cases},\\
        I_{\textrm{8fs}}&\begin{cases}
            \phi_{21} = I_{\textrm{8fs}}             & \phi_{22} = \exp{w_{22}I_{\textrm{8fs}}}-1 \\
            \phi_{23} = I_{\textrm{8fs}}^2            & \phi_{24} = \exp{w_{24}I_{\textrm{8fs}}^2}-1
        \end{cases},\\
        I_{\textrm{8fn}}&\begin{cases}
            \phi_{25} = I_{\textrm{8fn}}             & \phi_{26} = \exp{w_{26}I_{\textrm{8fn}}}-1 \\ 
            \phi_{27} = I_{\textrm{8fn}}^2            & \phi_{28} = \exp{w_{28}I_{\textrm{8fn}}^2}-1
        \end{cases},\\
        I_{\textrm{8sn}}&\begin{cases}
            \phi_{29} = I_{\textrm{8sn}}             & \phi_{30} = \exp{w_{30}I_{\textrm{8sn}}}-1 \\
            \phi_{31} = I_{\textrm{8sn}}^2            & \phi_{32} = \exp{w_{32}I_{\textrm{8sn}}^2}-1 
        \end{cases}.
    \end{aligned}
\end{equation}
The full vector of non-linear parameters is $\vecw = [w_{2}, w_{4}, \dots, w_{32}]\transpose \in \mathbb{R}^{16}$, while the linear coefficients are $\vecc \in \mathbb{R}^{32}$. However, as indicated in Section~\ref{subsec:anisotropic_library}, in the numerical test, we do not consider the linear and exponential-linear terms based on the corrected fourth invariants as well as the linear and exponential-linear terms based on the eighth invariants, since these terms can be different from zero in deformation-free state and thereby induce undesired residual stresses.

\newpage
\bibliographystyle{elsarticle-num} 
\bibliography{literature_R1}

\end{document}